\documentclass[letterpaper,10pt]{article}

\usepackage[utf8]{inputenc} 
\usepackage[T1]{fontenc}    
\usepackage{booktabs}       
\usepackage{amsfonts}       
\usepackage{nicefrac}       
\usepackage{microtype}      

\usepackage{subcaption}

\usepackage[table,xcdraw]{xcolor}

\usepackage{tikz}
\usetikzlibrary{fit}
\tikzset{%
  highlight/.style={rectangle,blend mode = multiply,draw=blue!90!black,thick,rounded corners = 0.3 mm,inner sep=0.5pt}
}

\usepackage{microtype}
\usepackage{graphicx}
\usepackage{booktabs} 

\usepackage{amsmath,amssymb,amsthm}
\usepackage{algorithm,algorithmicx,algpseudocode}
\usepackage{bbold}
\usepackage[sans]{dsfont}
\usepackage[numbers]{natbib}
\usepackage{cleveref}

\usepackage{pifont}

\usepackage{thmtools}
\usepackage{thm-restate}

\usepackage{multirow}

\usepackage{tabularx}

\usepackage{textcomp}

\usepackage[shortlabels]{enumitem}

\def\mat{\mathbf}
\def\wt{\widetilde}
\def\succeq{\succcurlyeq}
\def\Tr{\textnormal{Tr}}
\def\Ind#1{\mathbb{I}_{#1 \times #1}}
\def\st{\quad\textnormal{subject to}\quad}

\newcommand{\commentColor}{blue!90!black}

\let\oldComment\Comment
\renewcommand{\Comment}[1]{\oldComment{\textcolor{\commentColor}{#1}}}

\declaretheorem[name=Lemma,within=section]{lemma}
\declaretheorem[name=Corollary,within=section]{corollary}

\declaretheorem[name=Remark,within=section]{remark}
\declaretheorem[name=Definition,within=section]{definition}

\numberwithin{equation}{section}

\usepackage{wrapfig}

\usepackage[verbose=true,letterpaper]{geometry}
\AtBeginDocument{
  \newgeometry{
    textheight=9in,
    textwidth=6in,
    top=1in,
    headheight=12pt,
    headsep=25pt,
    footskip=30pt
  }
}

\usepackage{authblk}

\title{Learning Feature Sparse Principal Subspace}
\author{Lai Tian}
\author{Feiping Nie}
\author{Xuelong Li}
\affil{School of Computer Science, and \\
Center for OPTical IMagery Analysis and Learning (OPTIMAL),\\ Northwestern Polytechnical University, China. 
}
\date{\small\textsf{ Apr 23, 2019 } }

\begin{document}

\maketitle

\begin{abstract}
This paper presents new algorithms to solve the feature-sparsity constrained PCA problem (FSPCA), which performs feature selection and PCA simultaneously.
Existing optimization methods for FSPCA require data distribution assumptions and are lack of global convergence guarantee.
Though the general FSPCA problem is NP-hard, we show that, for a low-rank covariance, FSPCA can be solved globally (Algorithm \ref{alg:kISc}).
Then, we propose another strategy (Algorithm \ref{alg:generalFSPCA}) to solve FSPCA for the general covariance by iteratively building a carefully designed proxy. 
We prove theoretical guarantees on approximation and convergence for the new algorithms.
Experimental results show the promising performance of the new algorithms compared with the state-of-the-arts on both synthetic and real-world datasets.
\end{abstract}

\section{Introduction}

Consider $n$ data points in $\mathbb{R}^d$. When $d \gg n$, PCA has inconsistence issue in estimating the $m$ leading eigenvectors $\mat{W}\in\mathbb{R}^{d\times m}$ of population
covariance matrix $\mat{A} \in \mathbb{R}^{d\times d}$ \citep{johnstone2009consistency},
which can be addressed by assuming the sparsity in the principal components.
Prior work has been done in methodology design \citep{zou2006sparse,shen2008sparse,d2007direct,vu2013fantope,papailiopoulos2013sparse,yang2015streaming,kundu2017recovering} and theoretical understanding \citep{vu2013minimax,lei2015sparsistency,yang2016rate,zhang2018optimal}.

The principal subspace estimation problem is directly connected to dimension reduction and is important when there are more than one principal component of interest. 
Indeed, typical applications of PCA use the projection onto the principal subspace to facilitate exploration and inference of important features of the data. As the authors of \citep{vu2013minimax} point out, dimension reduction by PCA should emphasize subspaces rather than eigenvectors. The sparsity level in sparse principal subspace estimation is defined as follows \citep{vu2013minimax,wang2014tighten,yang2015streaming}.

\begin{definition}[Subspace sparsity, \citep{vu2013minimax}]\label{def:sparsespace}
For the $m$-dimensional principal subspace $\textnormal{span}(\mat{W})$ of the covariance $\mat{A}$, the subspace sparsity level $k$, which should be rotation invariant since the top $m$ eigenvalues of $\mat{A}$ might be not distinct, is defined by
\[
k = \textnormal{card}(\textnormal{supp}[\textnormal{diag}(\mat{\Pi})]) = \|\mat{W}\|_{2,0},
\]
where $\mat{\Pi} = \mat{WW}^\top$ is the projection matrix onto $\textnormal{span}(\mat{W})$.
\end{definition}

This paper considers the principal subspace estimation problem with the feature subspace sparsity constraint, termed Feature Sparse PCA (Problem \eqref{eq:fspca} ).
Some approaches have been proposed to solve the FSPCA problem \citep{wang2014tighten,yang2015streaming,magdon2016optimal}.
Yet, there are some drawbacks in the existing methods. 
(1) Most of the existing analysis only holds in high probability when specific data generation assumptions hold, e.g., \citet{yang2015streaming} requires data generated from the spike model, \citet{wang2014tighten} requires data generated from the Gaussian distribution. Otherwise, they only guarantee convergence when the initial solution is near the global optimum.
(2) Existing iterative schemes are not ascent guaranteed. 
(3) Some methods make the spike model assumption, in which the covariance is instinctively low-rank, but existing methods cannot make full use of the low-rank structure in the covariance.

\begin{figure}
 \centering
        \includegraphics{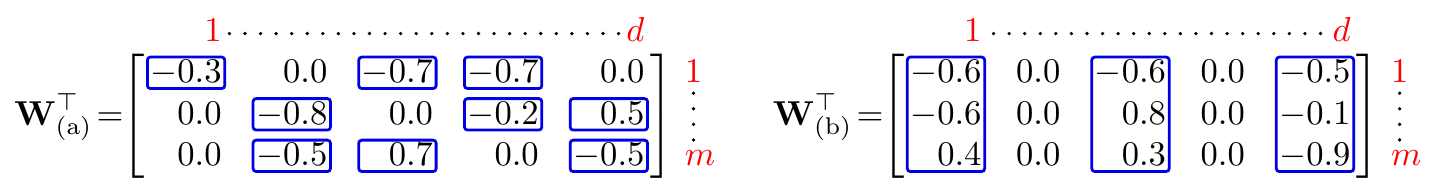}%
\caption{Element-wise Sparse PCA $\mat{W}_{\text{(a)}}$ versus Feature Sparse PCA $\mat{W}_{\text{(b)}}$.}%
\label{fig:matrix}%
\end{figure}
In this paper, we provide two optimization strategies to estimate the leading sparse principal subspace with provable guarantees.
The first strategy (Algorithm \ref{alg:kISc}) solves the feature sparse PCA problem globally when the covariance matrix is low-rank,
while the second strategy (Algorithm \ref{alg:generalFSPCA}) solves the feature sparse PCA for general covariance matrix iteratively with its convergence guaranteed.

\paragraph{Contributions.} More precisely, we make the following contributions:
\begin{enumerate}
  \item We show that, for a low-rank covariance matrix, the FSPCA problem can be solved globally with the newly proposed algorithm (Algorithm \ref{alg:kISc}). For the general high-rank case, we report an iterative algorithm (Algorithm \ref{alg:generalFSPCA}) by building a carefully designed proxy covariance.
  \item We prove theoretical guarantees on approximation and convergence for the proposed optimization strategies. Computational complexities analysis of both algorithms are provided.
  \item We conduct experiments on both synthetic and real-world data to evaluate the new algorithms. The experimental results demonstrate the promising performance of the newly proposed algorithms compared with the state-of-the-art methods.
\end{enumerate}

\paragraph{Notations.} Throughout this paper, scalars, vectors and matrices are denoted by lowercase letters, boldface lowercase letters and boldface uppercase letters, respectively; for a matrix $\mat{A} \in \mathbb{R}^{d \times d}$, $\mat{A}^\top$ denotes the transpose of $\mat{A}$, $\Tr(\mat{A}) = \sum_{i=1}^d a_{ii}$, $\| \mat{A} \|_F^2 = \Tr(\mat{A}^\top\mat{A})$; $\mathbb{1}_n\in\mathbb{R}^n$ denotes vector with all ones; $\|\mat{x}\|_0$ denotes the number of non-zero elements;$\|\mat{A}\|_{p,q} = ( \sum_{i=1}^d \|\mat{a}_i\|_p^q )^{1/q}$; $\Ind{n} \in \mathbb{R}^{n\times n}$ denotes the identity matrix; $\mat{H}_n = \Ind{n} - \frac{1}{n}\mathbb{1}_n\mathbb{1}_n^\top$ is the centralization matrix; $\mathcal{I}(1:k)$ is the first $k$ elements in indices $\mathcal{I}$; $\mat{A}^\dagger$ denotes the Moore–Penrose inverse; $\mat{A}_m$ is the best rank-$m$ approximation of $\mat{A}$ in Frobenius norm; card$(\mathcal{I})$ is the cardinality of $\mathcal{I}$; $\mathbb{1}$\{condition\} is the indicator of the condition.

\section{Prior Work}\label{sec:prior}

\paragraph{Sparse Principal Components.}
Most existing methods in the literature to solve the sparse PCA problem only estimate the first leading eigenvector with the element-wise sparsity constraint.
To estimate the $m$ leading eigenvectors, one has to build a new covariance matrix with the deflation technique \citep{mackey2009deflation} and solve the leading eigenvector again.
The main drawback of this scheme is that, for example, the indices of non-zero elements in the first eigenvector might not
be the same as that of the second eigenvector.  As shown in \Cref{fig:matrix}, the sparsity pattern is inconsistent among the $m$ leading eigenvectors. 
Moreover, as pointed out by \cite{wang2014tighten}, the deflation has identifiability and orthogonality issues when the top $m$ eigenvalues are not distinct. \citep{asteris2014nonnegative,papailiopoulos2013sparse} propose methods for the leading eigenvector with approximation guarantee but their guarantee only applies to the first component, not to further iterates.

\paragraph{Sparse Principal Subspace.}
\citet{vu2013minimax} consider a different setting that the estimated subspace is subspace sparsity constrained (\Cref{def:sparsespace}), in which the sparsity pattern is forced consistent among rows.
They show this problem has nice statistical properties \citep{vu2013minimax}, that is, the optimum is minimax optimal. But there is a gap between the computational method and statistical theory.
To close this gap, \citep{wang2014tighten,yang2015streaming,magdon2016optimal} proposed algorithms to solve the subspace sparsity constrained problem.
However, existing methods require data distribution assumptions and are lack of global convergence guarantee.

\paragraph{Sparse Regression.}
Another line of research \citep{pang2018efficient,du2018exploiting,cai2013exact} considers solving the sparse regression problem with the $\ell_{2,0}$ constraint. 
The main technical difference between the $\ell_{2,0}$ constrained sparse regression and FSPCA is the semi-orthogonal constraint on $\mat{W}$. Without the semi-orthogonal constraint, the FSPCA problem is not bound from above. 
Existing techniques to solve the $\ell_{2,0}$ constrained sparse regression problem, e.g., the projected gradient scheme in \cite{pang2018efficient}, cannot be used to solve our problem because, to our knowledge, there is no method to solve the projection subproblem with the semi-orthogonal constraint. Thus, the FSPCA problem is substantially more difficult than that of $\ell_{2,0}$-constrained sparse regression.
\section{Problem Setup}\label{sec:FSPCA}

Formally, we propose algorithms to solve the following general problem
\begin{equation}\label{eq:fspca}
\max_{\mat{W}\in\mathbb{R}^{d\times m}} \Tr\left( \mat{W}^\top \mat{A} \mat{W} \right) \st \mat{W}^\top\mat{W} = \Ind{m},\|\mat{W}\|_{2,0}\leq k,
\end{equation}
where $m \leq k \leq d$ and matrix $\mat{A} \in \mathbb{R}^{d\times d}$ is positive semi-definite.
This problem is NP-hard to solve globally even for $m = 1$ \citep{moghaddam2006spectral} and 
sadly NP-hard to solve $(1-\varepsilon)$-approximately for small $\varepsilon > 0$ \citep{chan2016approximability}.
Several techniques have been proposed \citep{wang2014tighten,yang2015streaming} to solve this challenging problem.
However, they only report high-probability analysis and none of them provides deterministic guarantee on both approximation and global convergence.

\begin{remark}
As shown in \citet{vu2013minimax}, the optimal $\mat{W}$ of Problem \eqref{eq:fspca} achieves the optimal minimax error for row sparse subspace estimation.
Besides, the FSPCA problem can be viewed as performing unsupervised feature selection and PCA simultaneously.
The key point is the $\ell_{2,0}$ norm constraint forces the sparsity pattern consistence among different eigenvectors,
while the vanilla element-wise sparse PCA model cannot keep this consistence as shown in \Cref{fig:matrix}.
One might use only the leading eigenvector for feature selection \citep{luss2010clustering,naikal2011informative} but this leads to suboptimal solution when there are more than one principal component of interest. 
\end{remark}
\section{Optimization Strategies}\label{sec:optimization}

In this section, we provide new optimization strategies to solve the FSPCA model in Problem \eqref{eq:fspca}.
We first consider the case when rank$(\mat{A}) \leq m$, for which a non-iterative strategy (Algorithm \ref{alg:kISc}) is provided to 
solve the problem globally. 
Then we consider the general case when rank$(\mat{A}) > m$, 
for which we provide an iterative algorithm (Algorithm \ref{alg:generalFSPCA}) by approximating $\mat{A}$ with a carefully designed low-rank proxy covariance $\mat{P}$  and solve the proxy subproblem with the Algorithm \ref{alg:kISc}.

\subsection{GO: Finding the Global Optimum for rank$(\mat{A}) \leq m$}

We make the following notion for ease of notations.
\begin{definition}[Row selection matrix map]
We define $(d, k)$-row selection matrix map $\mathbb{S}_{d,k}(\mathcal{I})$ to build row selection matrix $\mat{S}\in\mathbb{R}^{d\times k}$ according to given indices $\mathcal{I}$ such that $\mathbb{S}_{d,k}(\mathcal{I})=\mat{S}$. 
One can left multiply the row selection matrix $\mat{S}$ to select specific $k$ rows from $d$ inputs. Specifically,
\[
s_{ij}=
\left\{ \begin{array}{rcl}
         1 & \mbox{for} & i = \mathcal{I}(j) \\ 
         0 & \mbox{for} & otherwise. 
\end{array}
\right.
\]
\end{definition}

The algorithm to solve Problem \eqref{eq:fspca} is summarized in the following Algorithm \ref{alg:kISc}.
\begin{algorithm}[H]  
\caption{Solve Problem \eqref{eq:fspca} with $\textnormal{rank}(\mat{A}) \leq m$}  
\label{alg:kISc}
\begin{algorithmic}[1] 
\Procedure{Go}{$\mat{A},m,k,d$} \Comment{require $\mat{A}\succeq \mathbb{0}$, rank$(\mat{A}) \leq m$, and $m\leq k \leq d$.}
\State $\mathcal{I} \leftarrow$ \texttt{Sort}($\text{diag}(\mat{A})$, \texttt{\textquotesingle{}descending\textquotesingle}, \texttt{\textquotesingle{}output indices\textquotesingle}); \Comment{$O(d\log d)$.}
\State $\mat{S} \leftarrow \mathbb{S}_{d,k}(\texttt{Sort}(\mathcal{I}(1:k), \texttt{\textquotesingle{}ascending\textquotesingle}))$; \Comment{build selection matrix $\mat{S}$ from $\mathcal{I}$. $O(k \log k)$.}
\State $\mat{V} \leftarrow$ \texttt{Eigenvectors}($\mat{S}^\top\mat{AS}$, \texttt{\textquotesingle{}output $m$ leading eigenvectors\textquotesingle}); \Comment{$O(k^3)$.}
\State \Return $\mat{W} \leftarrow \mat{SV}$; \Comment{$O(km)$.}
\EndProcedure
\end{algorithmic}  
\end{algorithm} 
The following theorem justifies the \textbf{global optimality} of the output of Algorithm \ref{alg:kISc}:
\begin{restatable}{theorem}{globaloptimality}\label{thm:globaloptimality}
Suppose $\mat{A}\succeq \mathbb{0}$ and  rank$(\mat{A}) \leq m$. 
Let $\mat{W}=\textsc{Go}(\mat{A},m,k,d)$ with $m\leq k \leq d$. Then, $\mat{W}$ is a globally optimal solution of Problem \eqref{eq:fspca}.
\end{restatable}%
\begin{remark}\label{rem:lowRankApp}
\Cref{thm:globaloptimality} guarantees the global optimality of Algorithm \ref{alg:kISc} for a low-rank $\mat{A}$.
It is interesting to see that, though the Problem \eqref{eq:fspca} is NP-hard to solve in general, it is globally solvable 
for a low-rank covariance $\mat{A}$. A natural idea then comes out that we can try to solve the general Problem \eqref{eq:fspca} by running Algorithm \ref{alg:kISc} with the best rank-$m$ approximation $\mat{A}_m$. In \Cref{thm:approx} and \Cref{sec:exps}, we will justify this idea theoretically and empirically.
\end{remark}
\begin{remark}\label{rem:addEyeIsOK}
If there exists $\sigma$ such that $\textnormal{rank}(\mat{A}+\sigma\Ind{d}) \leq m$, which is the covariance in spike model, then the Algorithm \ref{alg:kISc} still outputs a globally optimal solution with $\mat{A} + \sigma\Ind{d}$ as input, since 
$
\Tr\left(\mat{W}^\top (\mat{A}+\sigma \Ind{d}) \mat{W} \right) = \Tr\left(\mat{W}^\top \mat{A} \mat{W} \right) + \sigma m.
$
$\textnormal{rank}(\mat{A}) \leq m$ is a special case when $\sigma=0$.
\end{remark}

\subsection{IPU: Iteratively Proxy Update for rank$(\mat{A}) > m$}
In this subsection, we consider the general case, that is, rank$(\mat{A}) > m$.
The main idea is that we try to build a proxy covariance, say $\mat{P}$, of original $\mat{A}$ such that $\textnormal{rank}(\mat{P}) \leq m, \mat{P} \succeq \mathbb{0}$. Then we can run Algorithm \ref{alg:kISc} with the low-rank proxy $\mat{P}$ to solve the original problem iteratively.

\textbf{Proxy Construction.} With careful design, given the estimate $\mat{W}_t$ from the $t$th iterative step, we define the matrix
\[
\mat{P}_t = \mat{A}\mat{W}_{t}(\mat{W}_t^\top \mat{A}\mat{W}_t)^{\dagger}\mat{W}_t^\top\mat{A}
\]
as the low-rank proxy matrix of original $\mat{A}$. Then, We solve Problem \ref{eq:fspca} with the proxy $\mat{P}_t$ rather than $\mat{A}$. Following claim verifies the sufficient conditions for $\mat{P}_t$
to be solvable with Algorithm \ref{alg:kISc}.

\begin{restatable}{claim}{proxyclaim}\label{chaim:PisSolvable}
For each $t\geq 1$, $\mat{W}_t^\top\mat{W}_t = \Ind{m}$, it holds $\textnormal{rank}(\mat{P}_t) \leq m$, and $ \mat{P}_t \succeq \mathbb{0}.$
\end{restatable}

\textbf{Indices Selection.} With the proxy matrix $\mat{P}_t$ in hand, a natural idea is to iteratively update $\mat{W}$ by solving the following problem with Algorithm \ref{alg:kISc}:
\begin{equation}\label{eq:WwtDEF}
\wt{\mat{W}}_{t+1} \leftarrow \textsc{Go}(\mat{P}_t, m, k,d).
\end{equation}
But we can further refine the $\wt{\mat{W}}_{t+1}$ by performing eigenvalue decomposition on original $\mat{A}$ rather than on the proxy covariance $\mat{P}_t$, which will accelerate the convergence.

\textbf{Eigenvectors Refinement.} Note that $\wt{\mat{W}}_{t+1}$ can be written as $\wt{\mat{W}}_{t+1} = \mat{S}_{t+1}\wt{\mat{V}}_{t+1}$, where $\mat{S}_{t+1}$ is the selection matrix and $\wt{\mat{V}}_{t+1}$ is the eigenvectors in the row support of $\wt{\mat{W}}_{t+1}$. 
Then, $\wt{\mat{W}}_{t+1}$ can be further refined by fixing the selection matrix $\mat{S}_{t+1}$ and updating the eigenvectors $\mat{V}_{t+1}$ with
\begin{equation}\label{eq:solveVclever}
\mat{V}_{t+1} \leftarrow \mathop{\arg\max}_{\mat{V}^\top\mat{V}=\Ind{m}} \Tr(\mat{V}^\top\mat{S}_{t+1}^\top\mat{A}\mat{S}_{t+1}\mat{V}).
\end{equation}
And finally, the refined $\mat{W}_{t+1}$ can be computed by
$
\mat{W}_{t+1} \leftarrow \mat{S}_{t+1}\mat{V}_{t+1}.
$ 
Compared with updating with Problem \eqref{eq:WwtDEF}, updating with the refinement makes larger progress thus it is more aggressive.

In summary, we collect the proposed procedure to  solve FSPCA when $\textnormal{rank}(\mat{A}) > m$ in Algorithm \ref{alg:generalFSPCA}.

\begin{algorithm}[H]  
\caption{Solve Problem \eqref{eq:fspca} with $\textnormal{rank}(\mat{A}) > m$}  
\label{alg:generalFSPCA}
\begin{algorithmic}[1]  
\Procedure{Ipu}{$\mat{A},m,k,d, \mat{W}_0$} \Comment{require $\mat{A}\succeq \mathbb{0}$ and $m\leq k \leq d$.}
\Repeat
\State $\mat{P}_t \leftarrow \mat{A}\mat{W}_t(\mat{W}_t^\top \mat{A}\mat{W}_t )^{\dagger}\mat{W}_t^\top \mat{A}$; \Comment{$O(m^2d)$.}
\State $\mathcal{I}_t \leftarrow$ \texttt{Sort}($\text{diag}(\mat{P}_t)$, \texttt{\textquotesingle{}descending\textquotesingle{}}, \texttt{\textquotesingle{}output indices\textquotesingle{}}); \Comment{$O(d\log d)$.}
\State $\mat{S}_t \leftarrow \mathbb{S}_{d,k}(\texttt{Sort}(\mathcal{I}_t(1:k), \texttt{\textquotesingle{}ascending\textquotesingle}))$; \Comment{$O(k\log k)$.}
\State $\mat{V}_t \leftarrow$ \texttt{Eigenvectors}($\mat{S}^\top\mat{AS}$, \texttt{\textquotesingle{}output $m$ leading eigenvectors\textquotesingle{}}); \Comment{$O(k^3)$.}
\State 
$
\mat{W}_{t+1} \leftarrow \mat{S}_{t}\mat{V}_{t}
$; $\quad t \leftarrow t+1$; \label{alg_line:defW} \Comment{$O(km)$.}
\Until{$\mathcal{I}_{t-1} = \mathcal{I}_{t-2}$} 
\State \Return{$\mat{W}_t$;}
\EndProcedure
\end{algorithmic}  
\end{algorithm} 

\section{Theoretical Justification}\label{sec:analysis}

In this section, we provide the theoretical analysis for Algorithm \ref{alg:kISc} and \ref{alg:generalFSPCA}. 
In detail, we prove approximation and convergence guarantees for the new algorithms. Then, we report the computational complexities and compare them with these of methods in the literature.

\subsection{Approximation Guarantee}\label{sec:approx}
First, we define constants related to the eigenvalues decay of $\mat{A}$. Let $r=\min\{\textnormal{rank}(\mat{A}), 2m\}$, and
\[\textstyle
G_1 = \frac{\sum_{i=m+1}^{r}\lambda_i(\mat{A})}{\sum_{i=1}^m \lambda_i(\mat{A})},\qquad\qquad
G_2 = \frac{\sum_{i=m+1}^{r}\lambda_i(\mat{A})}{\sum_{i=1}^d \lambda_i(\mat{A})}.
\]
The main approximation result can be stated as follows.
\begin{restatable}{theorem}{approxgua}\label{thm:approx}
Suppose $\mat{A}\succeq \mathbb{0}, m\leq k \leq d$. Let $\mat{W}_m = \text{Go}(\mat{A}_m, m, k, d)$, and $\mat{W}_*$ be a globally optimal solution of Problem \ref{eq:fspca}.
Then, we have
$
(1-\varepsilon) \leq \frac{\Tr(\mat{W}_m^\top\mat{A}\mat{W}_m) }{\Tr(\mat{W}_*^\top\mat{A}\mat{W}_*)} \leq 1
$
with $\varepsilon \leq \min\left\{ \frac{dG_1}{k},\frac{dG_2}{m} \right\}$.
\end{restatable}

\begin{remark}
\Cref{thm:approx} says that, for sufficiently large $m$ or $k$, $\text{Go}(\mat{A}_m, m, k, d)$ gives a $(1-\varepsilon)$-approximate solution of Problem \ref{eq:fspca}. Also note that, when the eigenvalues of the covariance $\mat{A}$ decay fast enough, a small $m$ or $k$ is sufficient to guarantee $(1-\varepsilon)$-approximation.
It is notable that, when $\textnormal{rank}(\mat{A}) \leq m$, we have $G_1 = G_2 = 0, \mat{A}=\mat{A}_m$, which implies $\varepsilon=0$ and the output of the Algorithm \ref{alg:kISc} is a globally optimal solution. This cross-verifies the correctness of \Cref{thm:globaloptimality}.
\end{remark}

It has been observed by \citet{breslau1999web,faloutsos1999power,mihail2002eigenvalue} that many phenomena approximately follow Zipf-like spectrum, e.g., Web caching, Internet topology, and city population. Specifically, the $i$th eigenvalue of the Zipf-like spectrum is $ci^{-t}$ with constants $c>0,t>1$.
We have following corollary from \Cref{thm:approx} for Zipf-like distributed eigenvalues.

\begin{restatable}[Zipf-like distribution]{corollary}{coroPower}\label{coro:power}
Suppose $\mat{A}\succeq \mathbb{0}, m\leq k \leq d$, and $\lambda_i(\mat{A})=ci^{-t}$ with $t > 1, c > 0$ for each $i=1,\dots,2m$. 
Let $\mat{W}_m = \text{Go}(\mat{A}_m, m, k, d)$, and $\mat{W}_*$ be an optimal solution of Problem \ref{eq:fspca}. If the sparsity satisfies $k=\Omega\left(\frac{d}{\varepsilon m^{t-1}}\right)$, then we have
$(1-\varepsilon) \leq \frac{\Tr(\mat{W}_m^\top\mat{A}\mat{W}_m)}{\Tr(\mat{W}_*^\top\mat{A}\mat{W}_*)} \leq 1$.
\end{restatable}

\subsection{Convergence Guarantee}

In this section, we show the iterative scheme proposed in Algorithm \ref{alg:generalFSPCA} increases the objective function value in every iterative step, which directly indicates the convergence of the iterative scheme.

\begin{restatable}{theorem}{ascentthm}\label{thm:ascent}
Suppose $\mat{A}\succeq \mathbb{0}, m\leq k \leq d$. Let $\mat{W}_{t+1}=\mat{S}_t\mat{V}_t$ be the variable defined in \Cref{alg_line:defW} of Algorithm \ref{alg:generalFSPCA}. Then, it holds that
$
\Tr(\mat{W}_{t}^\top\mat{A}\mat{W}_{t})\leq \Tr(\mat{W}_{t+1}^\top\mat{A}\mat{W}_{t+1}).
$
\end{restatable}
\begin{remark}\label{rem:ascent}
\Cref{thm:ascent} shows that the newly proposed Algorithm \ref{alg:generalFSPCA} is an ascent method, that is $\{\Tr(\mat{W}_t^\top \mat{A}\mat{W}_t)\}_{t=1}^T$ is an increasing sequence, which is important since most of the existing algorithms for solving Problem \eqref{eq:fspca} are not ascent.  That is to say, they cannot guarantee the output is better than the initialization.
Combining with the fact that the objective function is bounded from above by finite $\Tr(\mat{A})$, the convergence of Algorithm \ref{alg:generalFSPCA} can be obtained.
\end{remark}

Leveraging the ascent property, we have following the approximation guarantee for Algorithm \ref{alg:generalFSPCA}.
\begin{corollary}\label{coro:approxIPU}
Suppose $\mat{A}\succeq \mathbb{0}$. Let $\widehat{\mat{W}} = \text{IPU}(\mat{A}, m, k, d, \text{Go}(\mat{A}_m, m, k, d))$, and $\mat{W}_*$ be an optimal solution of Problem \ref{eq:fspca}.
Then, we have
$
(1-\varepsilon) \leq \frac{\Tr(\widehat{\mat{W}}^\top\mat{A}\widehat{\mat{W}}) }{\Tr(\mat{W}_*^\top\mat{A}\mat{W}_*)} \leq 1
$
with $\varepsilon \leq \min\left\{ \frac{dG_1}{k},\frac{dG_2}{m} \right\}$.
\end{corollary}

\subsection{Computational Complexity}

\textbf{Algorithm \ref{alg:kISc}.} It is easy to see the overall complexity is $O(
\max\{d\log d, k^3\})$ since $O(d\log d)$ for indices selection, $O(k^3)$ for eigenvalue decomposition, and $O(km)$ for building the output $\mat{W}$. 

\textbf{Algorithm \ref{alg:generalFSPCA}.} The overall computational complexity is $O(
\max\{d\log d, k^3, dkm\}T)$, where $T$ is the number of iterative steps used to coverage.
In \Cref{sec:toydata}, we will see the number of iterative steps $T$ is usually less than $20$ empirically.
For proxy covariance construction and indices selection, we need $O(\max\{dm^2, d\log d\})$ for naively building $\mat{P}_t$ and running Algorithm \ref{alg:kISc}. But note that we only need the diagonal elements in $\mat{P}_t$ for sorting and selecting.
Thus, we only compute the diagonal elements of $\mat{P}_t$ and sort it for the indices selection, that is $O(\max\{dkm, d\log d\})$. Then, performing eigenvectors refinement and updating $\mat{W}_{t+1}$ costs $O(k^3)$.
Also note that, the computational complexity of SOAP proposed in \citep{wang2014tighten} is $O(d^2m)$ for every iterative step. Ours computational complexity is strictly less than that of SOAP. For SRT in \citep{yang2015streaming}, the computational complexity is $O(dm \min\{m, k \log d\}) $. When $k=O(m)$, our complexity matches that of SRT.

\section{Discussion}\label{sec:discuss}

In this section, we provide discussion to show that the newly proposed algorithm fits into the MM optimization framework and discuss the invertibility issue of $\mat{W}_t^\top\mat{A}\mat{W}_t$.

\subsection{MM Framework}
Lots of classical algorithms can be framed into the MM framework, e.g., EM Algorithm \citep{dempster1977maximum}, Proximal Algorithms \citep{bertsekas1994partial,parikh2014proximal}, Concave-Convex Procedure (CCCP) \citep{yuille2002concave,lipp2016variations}.
Please refer to \citep{sun2017majorization} for further discussion.
It is notable that the newly proposed Algorithm \ref{alg:generalFSPCA} can also be viewed as a special case of the general MM optimization framework with the auxiliary function defined by
$
g(\mat{W};\mat{W}_t) = \Tr(\mat{W}^\top \mat{A}\mat{W}_{t}(\mat{W}_t^\top \mat{A}\mat{W}_t)^{\dagger}\mat{W}_t^\top\mat{A} \mat{W})
\leq  \Tr(\mat{W}^\top \mat{A} \mat{W}).
$
Meanwhile, it is easy to check that $g(\mat{W};\mat{W}_t)$ satisfies
$
g(\mat{W}_t;\mat{W}_t) = \Tr(\mat{W}_t^\top \mat{A} \mat{W}_t).
$
\subsection{On the Invertibility of $\mat{W}_t^\top\mat{A}\mat{W}_t$}
In the definition of the proxy matrix $\mat{P}_t$, there is a Moore–Penrose inverse term $(\mat{W}_t^\top\mat{A}\mat{W}_t)^\dagger$.
In this subsection we provide a condition under which this matrix is always invertible thus the Moore–Penrose inverse can be replaced with the matrix inverse. 
The reason why we care about the invertibility is that when $\mat{W}_t^\top\mat{A}\mat{W}_t$ is not invertible, it is rank deficient.
Thus it might not be a good approximation to the high-rank covariance $\mat{A}$. 

\begin{restatable}{claim}{invertclaim}\label{claim:invert}
If \textnormal{rank}$(\mat{A}) \geq d - k + m$, then, for all $t$, $\mat{W}_t^\top\mat{A}\mat{W}_t$ in Algorithm \ref{alg:generalFSPCA} is always invertible.
\end{restatable}

\begin{remark} \label{rem:invert}
Note that the condition shown in \Cref{claim:invert} is easy to be satisfied. Indeed, we can solve Problem \eqref{eq:fspca} with $\mat{A}_\varepsilon = \mat{A} + \varepsilon\cdot\Ind{d}$.
Thus, $\textnormal{rank}(\mat{A}_\varepsilon) = d \geq d-k+m$.
Note that this small $\varepsilon$ perturbation on $\mat{A}$ does not change the optimal $\mat{W}$ because
$
\Tr\left(\mat{W}^\top \mat{A}_\varepsilon \mat{W} \right) = \Tr\left(\mat{W}^\top \mat{A} \mat{W} \right) + \varepsilon m,
$
which is only a constant $\varepsilon m$ added to the original objective function. Thus, the optimal $\mat{W}$ remains unchanged.
In practice, we recommend using $\mat{A}_\varepsilon$ with a small $\varepsilon > 0$ to keep safe.
\end{remark}

\begin{table}[t]
\centering
\caption{Synthetic Data Description}
\label{tab:my-table}
\begin{tabularx}{\textwidth}{@{}clXl@{}}
\toprule
No. & \multicolumn{1}{c}{Description}                                                      &  & \multicolumn{1}{c}{Note}                  \\ \midrule
A   & $\mat{\lambda}(\mat{A})=\{100,100,4,1,\dots,1\}$                                     &  & Setting in \citep{wang2014tighten}           \\
B   & $\mat{\lambda}(\mat{A})=\{300,180,60,1,\dots,1\}$                                    &  & Setting in \citep{wang2014tighten}           \\
C   & $\mat{\lambda}(\mat{A})=\{300,180,60,0,\dots,0\}$                                    &  & To justify \Cref{thm:globaloptimality}       \\
D   & $\mat{\lambda}(\mat{A})=\{160,80, 40, 20, 10, 5, 2,1,\dots,1\}$                      &  & For all $\sigma$, rank$(\mat{A+\sigma\Ind{d}}) > m$ \\
E   & $\mat{X}$ is \emph{iid} sampled from $\mathcal{U}[0, 1]$ and $\mat{A}=\mat{XX}^\top$ &  & Uniform Distribution                      \\
F   & $\mat{X}$ is \emph{iid} sampled from $\mathcal{N}(0, 1)$ and $\mat{A}=\mat{XX}^\top$ &  & Gaussian Distribution                     \\ \bottomrule
\end{tabularx}
\end{table}

\section{Experiments}\label{sec:exps}

In this section, we provide experimental results to validate the effectiveness of the proposed Go and IPU on both synthetic 
and real-world data. In our experiments, we always use $\mat{A}_\varepsilon$ with $\varepsilon=0.1$ to keep safe (\Cref{rem:invert}), except in the No. C synthetic data where we require $\textnormal{rank}(\mat{A}) \leq m$.

\begin{table}[t]
\centering
\caption{Synthetic Data Results. [mean $($std. err.$)$; $\uparrow$: larger is better; $\downarrow$: smaller is better]}
\label{tab:toy}
\resizebox{\textwidth}{!}{%
\begin{tabular}{@{}ccccccccccccc@{}}
\toprule
                   &      & \multicolumn{3}{c}{Random Subspace}                                &           & \multicolumn{3}{c}{Convex Relaxation}                              &           & \multicolumn{3}{c}{Low Rank Approx.}                               \\ \cmidrule(lr){3-5} \cmidrule(lr){7-9} \cmidrule(l){11-13} 
                   &      & IR $\uparrow$        & RE $\downarrow$      & HF $\uparrow$        &           & IR $\uparrow$        & RE $\downarrow$      & HF $\uparrow$        &           & IR $\uparrow$        & RE $\downarrow$      & HF $\uparrow$        \\ \midrule
\multirow{5}{*}{A} & SOAP & 0.73 (0.09)          & 0.03 (0.02)          & 0.18 (0.15)          &           & 0.71 (0.12)          & 0.08 (0.04)          & 0.01 (0.01)          &           & 0.84 (0.12)          & 0.03 (0.03)          & 0.22 (0.17)          \\
                   & SRT  & 0.77 (0.19)          & 0.01 (0.02)          & 0.70 (0.21)          &           & 0.92 (0.12)          & 0.01 (0.02)          & 0.62 (0.24)          &           & 0.88 (0.16)          & 0.02 (0.04)          & 0.50 (0.25)          \\
                   & CSSP & 0.63 (0.13)          & 0.88 (0.05)          & 0.00 (0.00)          &           & 0.62 (0.12)          & 0.87 (0.06)          & 0.00 (0.00)          &           & 0.62 (0.12)          & 0.87 (0.06)          & 0.00 (0.00)          \\
                   & Go   & 0.92 (0.12)          & 0.01 (0.03)          & 0.74 (0.19)          &           & 0.93 (0.12)          & 0.01 (0.03)          & 0.67 (0.22)          &           & 0.93 (0.12)          & 0.01 (0.03)          & 0.66 (0.22)          \\
                   & IPU  & \textbf{0.97 (0.04)} & \textbf{0.00 (0.00)} & \textbf{1.00 (0.00)} & \textbf{} & \textbf{0.99 (0.04)} & \textbf{0.00 (0.00)} & \textbf{0.97 (0.03)} & \textbf{} & \textbf{0.98 (0.05)} & \textbf{0.00 (0.00)} & \textbf{0.91 (0.08)} \\ \midrule
\multirow{5}{*}{B} & SOAP & 0.76 (0.12)          & 0.03 (0.03)          & 0.14 (0.12)          &           & 0.78 (0.11)          & 0.04 (0.03)          & 0.09 (0.08)          &           & 0.77 (0.12)          & 0.04 (0.03)          & 0.05 (0.05)          \\
                   & SRT  & 0.59 (0.08)          & 0.03 (0.03)          & 0.28 (0.20)          &           & 0.79 (0.14)          & 0.04 (0.04)          & 0.15 (0.13)          &           & 0.80 (0.16)          & 0.04 (0.05)          & 0.30 (0.21)          \\
                   & CSSP & 0.77 (0.10)          & 0.90 (0.05)          & 0.00 (0.00)          &           & 0.76 (0.12)          & 0.90 (0.05)          & 0.00 (0.00)          &           & 0.76 (0.12)          & 0.91 (0.05)          & 0.00 (0.00)          \\
                   & Go   & \textbf{0.99 (0.02)} & \textbf{0.00 (0.00)} & \textbf{1.00 (0.00)} & \textbf{} & \textbf{0.99 (0.02)} & \textbf{0.00 (0.00)} & \textbf{1.00 (0.00)} & \textbf{} & \textbf{0.99 (0.01)} & \textbf{0.00 (0.00)} & \textbf{1.00 (0.00)} \\
                   & IPU  & 0.97 (0.03)          & \textbf{0.00 (0.00)} & \textbf{1.00 (0.00)} & \textbf{} & \textbf{0.99 (0.01)} & \textbf{0.00 (0.00)} & \textbf{1.00 (0.00)} & \textbf{} & \textbf{0.99 (0.01)} & \textbf{0.00 (0.00)} & \textbf{1.00 (0.00)} \\ \midrule
\multirow{5}{*}{C} & SOAP & 0.77 (0.12)          & 0.04 (0.03)          & 0.11 (0.10)          &           & 0.77 (0.12)          & 0.04 (0.03)          & 0.08 (0.07)          &           & 0.76 (0.12)          & 0.04 (0.03)          & 0.05 (0.05)          \\
                   & SRT  & 0.59 (0.08)          & 0.03 (0.04)          & 0.20 (0.16)          &           & 0.76 (0.16)          & 0.05 (0.05)          & 0.12 (0.11)          &           & 0.80 (0.17)          & 0.05 (0.06)          & 0.26 (0.19)          \\
                   & CSSP & 0.77 (0.11)          & 0.94 (0.03)          & 0.00 (0.00)          &           & 0.76 (0.12)          & 0.94 (0.03)          & 0.00 (0.00)          &           & 0.76 (0.12)          & 0.94 (0.03)          & 0.00 (0.00)          \\
                   & Go   & \textbf{1.00 (0.00)} & \textbf{0.00 (0.00)} & \textbf{1.00 (0.00)} & \textbf{} & \textbf{1.00 (0.00)} & \textbf{0.00 (0.00)} & \textbf{1.00 (0.00)} & \textbf{} & \textbf{1.00 (0.00)} & \textbf{0.00 (0.00)} & \textbf{1.00 (0.00)} \\
                   & IPU  & \textbf{1.00 (0.00)} & \textbf{0.00 (0.00)} & \textbf{1.00 (0.00)} & \textbf{} & \textbf{1.00 (0.00)} & \textbf{0.00 (0.00)} & \textbf{1.00 (0.00)} & \textbf{} & \textbf{1.00 (0.00)} & \textbf{0.00 (0.00)} & \textbf{1.00 (0.00)} \\ \midrule
\multirow{5}{*}{D} & SOAP & 0.79 (0.08)          & 0.01 (0.01)          & 0.43 (0.25)          &           & 0.80 (0.13)          & 0.02 (0.02)          & 0.15 (0.13)          &           & 0.84 (0.11)          & 0.01 (0.01)          & 0.22 (0.17)          \\
                   & SRT  & 0.57 (0.07)          & 0.02 (0.02)          & 0.14 (0.12)          &           & 0.77 (0.15)          & 0.04 (0.04)          & 0.12 (0.11)          &           & 0.83 (0.14)          & 0.02 (0.03)          & 0.27 (0.20)          \\
                   & CSSP & 0.76 (0.12)          & 0.80 (0.07)          & 0.00 (0.00)          &           & 0.77 (0.12)          & 0.82 (0.08)          & 0.00 (0.00)          &           & 0.77 (0.12)          & 0.82 (0.08)          & 0.00 (0.00)          \\
                   & Go   & \textbf{0.91 (0.10)} & \textbf{0.00 (0.01)} & 0.52 (0.25)          &           & 0.92 (0.10)          & \textbf{0.00 (0.01)} & 0.59 (0.24)          &           & \textbf{0.92 (0.09)} & \textbf{0.00 (0.01)} & 0.56 (0.25)          \\
                   & IPU  & 0.83 (0.07)          & \textbf{0.00 (0.00)} & \textbf{0.97 (0.03)} & \textbf{} & \textbf{0.93 (0.10)} & \textbf{0.00 (0.01)} & \textbf{0.65 (0.23)} & \textbf{} & \textbf{0.92 (0.10)} & \textbf{0.00 (0.01)} & \textbf{0.60 (0.24)} \\ \midrule
\multirow{5}{*}{E} & SOAP & 0.43 (0.07)          & 0.06 (0.03)          & 0.01 (0.01)          &           & 0.46 (0.16)          & 0.12 (0.05)          & 0.00 (0.00)          &           & 0.73 (0.16)          & 0.04 (0.04)          & 0.12 (0.11)          \\
                   & SRT  & 0.86 (0.07)          & \textbf{0.00 (0.00)} & 0.72 (0.20)          &           & 0.88 (0.11)          & 0.01 (0.01)          & 0.40 (0.24)          &           & \textbf{0.90 (0.09)} & \textbf{0.01 (0.01)} & \textbf{0.52 (0.25)} \\
                   & CSSP & 0.43 (0.16)          & 0.82 (0.06)          & 0.00 (0.00)          &           & 0.43 (0.16)          & 0.83 (0.06)          & 0.00 (0.00)          &           & 0.44 (0.16)          & 0.83 (0.06)          & 0.00 (0.00)          \\
                   & Go   & \textbf{0.89 (0.09)} & \textbf{0.00 (0.01)} & 0.48 (0.25)          &           & \textbf{0.90 (0.09)} & \textbf{0.00 (0.01)} & \textbf{0.46 (0.25)} &           & 0.88 (0.10)          & \textbf{0.01(0.01)}  & 0.41 (0.24)          \\
                   & IPU  & 0.83 (0.06)          & \textbf{0.00 (0.00)} & \textbf{0.89 (0.10)} &           & 0.87 (0.10)          & 0.01 (0.01)          & 0.37 (0.23)          &           & 0.88 (0.10)          & \textbf{0.01(0.01)}  & 0.42 (0.24)          \\ \midrule
\multirow{5}{*}{F} & SOAP & 0.61 (0.07)          & \textbf{0.01 (0.01)} & 0.36 (0.23)          &           & 0.79 (0.14)          & \textbf{0.03 (0.03)} & 0.16 (0.13)          &           & 0.81 (0.12)          & \textbf{0.03 (0.02)} & 0.16 (0.13)          \\
                   & SRT  & 0.62 (0.08)          & \textbf{0.01 (0.01)} & 0.37 (0.23)          &           & \textbf{0.82 (0.12)} & \textbf{0.03 (0.02)} & \textbf{0.20 (0.16)} &           & \textbf{0.82 (0.12)} & \textbf{0.03 (0.02)} & \textbf{0.17 (0.14)} \\
                   & CSSP & 0.79 (0.13)          & 0.52 (0.08)          & 0.00 (0.00)          &           & 0.77 (0.14)          & 0.54 (0.08)          & 0.00 (0.00)          &           & 0.77 (0.14)          & 0.54 (0.08)          & 0.00 (0.00)          \\
                   & Go   & \textbf{0.83 (0.12)} & 0.02 (0.03)          & 0.21 (0.17)          &           & 0.81 (0.12)          & \textbf{0.03 (0.03)} & 0.16 (0.13)          &           & 0.81 (0.12)          & \textbf{0.03 (0.03)} & 0.16 (0.13)          \\
                   & IPU  & 0.62 (0.07)          & \textbf{0.01 (0.01)} & \textbf{0.44 (0.25)} &           & \textbf{0.82 (0.12)} & \textbf{0.03 (0.02)} & 0.18 (0.15)          &           & \textbf{0.82 (0.12)} & \textbf{0.03 (0.02)} & \textbf{0.17 (0.14)} \\ \bottomrule
\end{tabular}%
}
\end{table}

\subsection{Synthetic Data}\label{sec:toydata}
To show the effectiveness of the proposed method, we build a series of small-scale synthetic datasets, whose global optimum can be obtained by brute-force searching. Then we compare  our methods with several state-of-the-art methods with optimal indices and objective value in hand.

\paragraph{Experiments Setup.} We compare the newly proposed \textsc{Go} (Algorithm \ref{alg:kISc}) 
and \textsc{Ipu} (Algorithm \ref{alg:generalFSPCA}) with SOAP \citep{wang2014tighten}, SRT \citep{yang2015streaming}, and 
CSSP \citep{magdon2016optimal}.
For the synthetic data, we fix $m=3,k=7,$and $d=20$. 
We cannot afford large-scale setting since the brute-force searching space grows exponentially.
We consider three different initialization methods: Random Subspace; Convex Relaxation proposed in \citep{vu2013fantope} and used in \citep{wang2014tighten}; Low Rank Approx. with \textsc{Go}$(\mat{A}_m, m, k, d)$. 
We consider $6$ different synthetic data in our experiments. The descriptions of these schemes are summarized in \Cref{tab:my-table}.
For A--D, we fix the eigenvalues and generate the eigenspace randomly.
Every scheme is independently run for 100 times and we report the mean and standard error. 
For the Random Subspace setting, every realization $\mat{A}$ is repeated run 20 times with different random initialization.
Thus, in the random initialization setting, we run all algorithms $20\times 100=2000$ times.
The overall mean and standard error are reported.

\paragraph{Performance Measures.} 
Intersection Ratio (IR): 
  $
\frac{\text{card}\left( \{\text{estimated indices}\} \cap \{\text{optimal indices} \} \right)}{\text{\# sparsity }k}.
  $
  Relative Error (RE):
  $
\frac{\Tr\left(\mat{W}^\top\mat{A}\mat{W}\right) - \Tr\left(\mat{W}_*^{\top}\mat{A}\mat{W}_*\right)}{\Tr\left(\mat{W}_*^{\top}\mat{A}\mat{W}_*\right)}.
  $
  Hit Frequency (HF):
  $
\frac{1}{N} \sum_{i=1}^N \mathbb{1}\{ \text{Relative Error} \leq 10^{-3}\},
  $%
  where $N$ is the number of repeated running. 
The motivation to use these measures is deferred to \Cref{sec:expDetail}.

\paragraph{Results.}
 Experimental results are reported in \Cref{tab:toy}, and  we get the following insights:
  (1) From No. C, Algorithm \ref{alg:kISc} gives a globally optimal solution when the covariance $\mat{A}$ is low-rank.
  (2) Both the performance of Go and IPU outperform or match other state-of-the-art methods, especially when the numerical rank of covariance is small.
  (3) CSSP does not perform well in HF and RE, which is consistent with results reported in \citep{magdon2016optimal}, since the objective of CSSP is a regression-type minimization rather than variance maximization.
  (4) When the Low Rank Approx. strategy (with Go) is used as initialization, all methods have match or even better explained variance than initialization with Convex Relaxation, while the computational complexity of Low Rank Approx. (with SVD) is seriously smaller than that of Convex Relaxation (with ADMM or SDP). Meanwhile, initialization with Random Space has better performance than both Convex Relaxation and Low Rank Approx., which is not surprising since the reported results for Random Subspace are the maximal objective value among 20 random initialization. This strategy is widely used in practice, e.g., run $k$-means multiple times with different initialization and pick the one with the smallest loss.

\subsection{Real-world Data}

\begin{figure}[t!]
    \centering
    \begin{subfigure}[t]{0.32\textwidth}
        \centering
        \includegraphics[width=\textwidth]{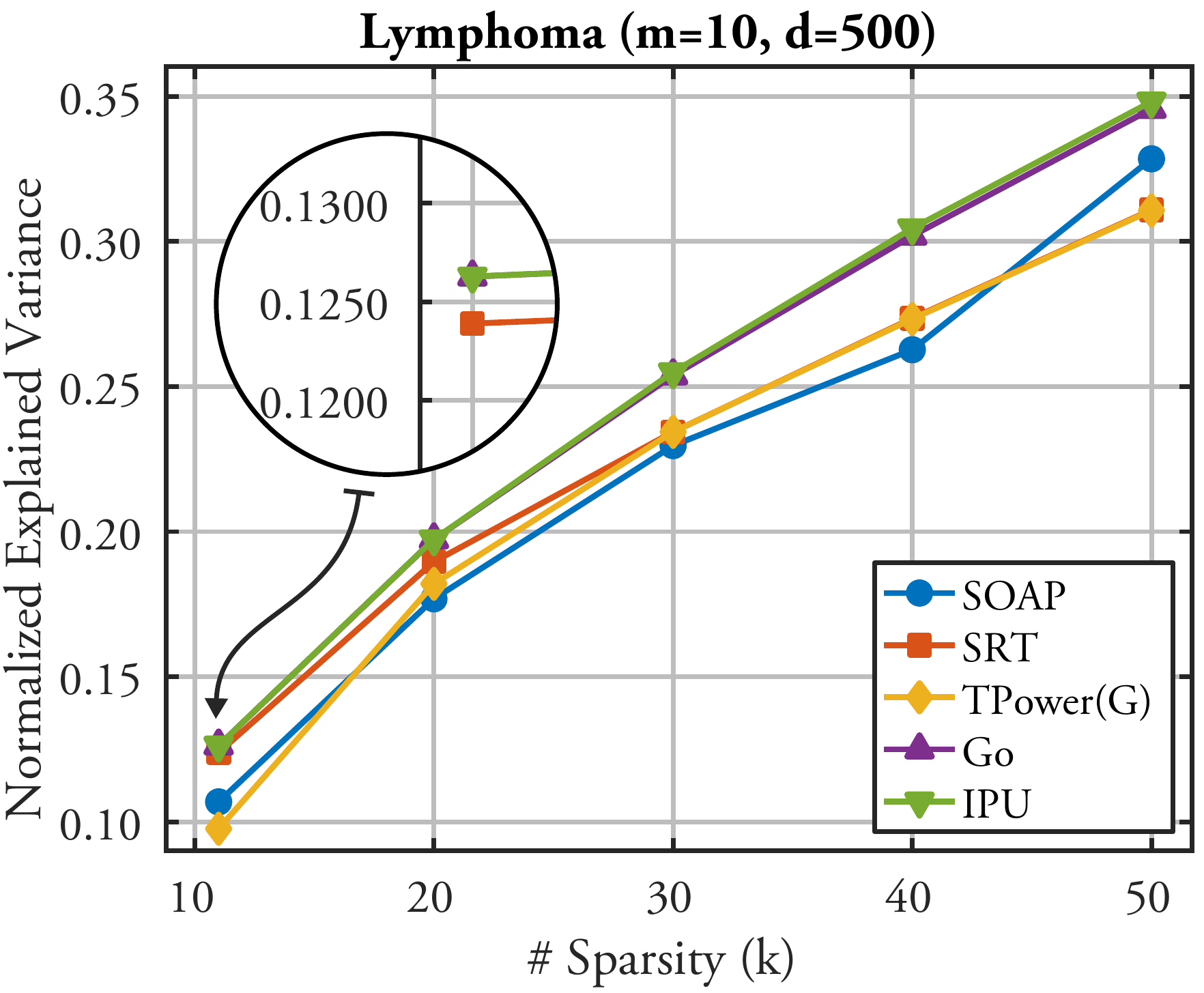}%
    \end{subfigure}%
    ~ 
    \begin{subfigure}[t]{0.32\textwidth}
        \centering
        \includegraphics[width=\textwidth]{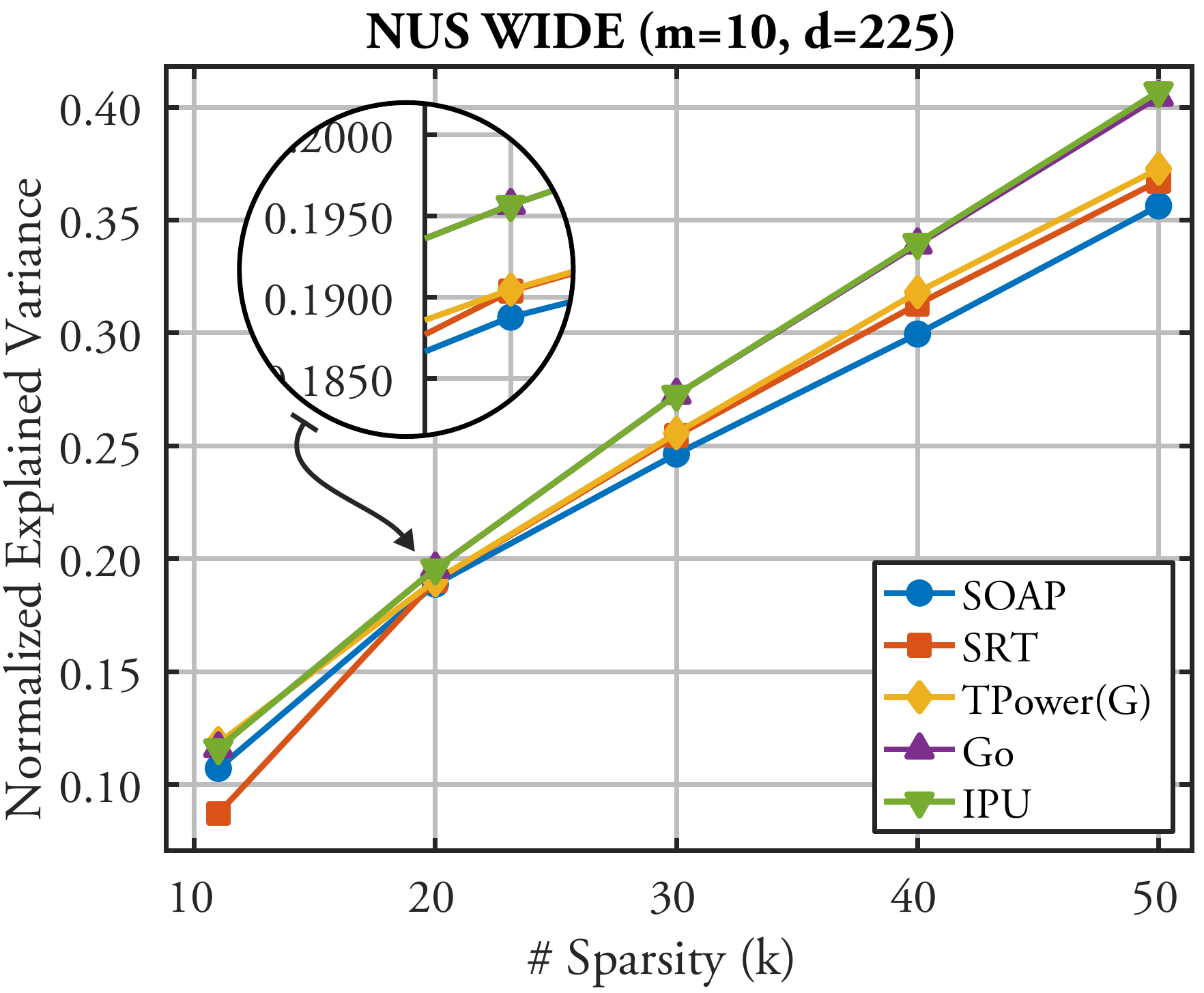}%
    \end{subfigure}
    ~ 
    \begin{subfigure}[t]{0.32\textwidth}
        \centering
        \includegraphics[width=\textwidth]{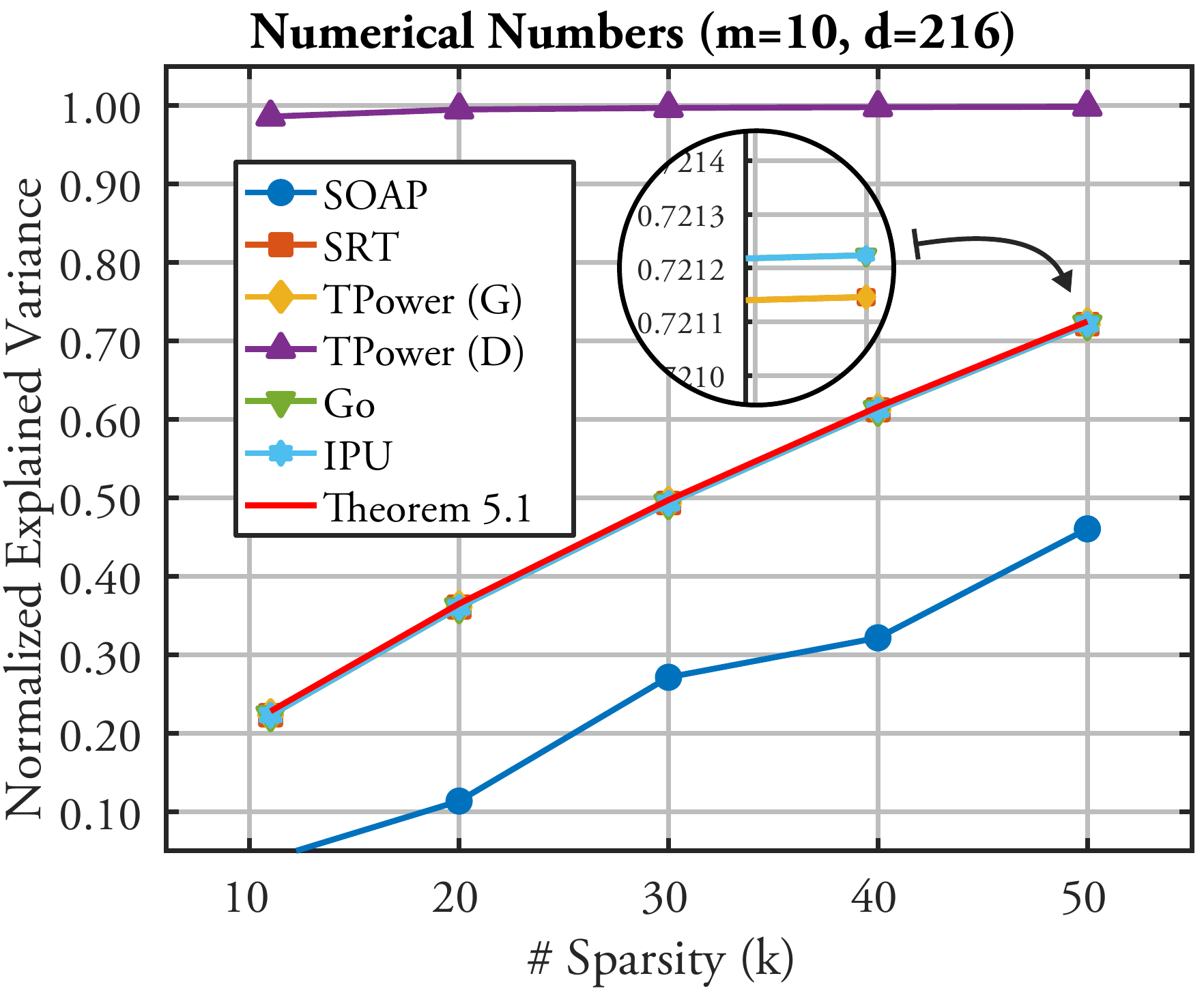}%
    \end{subfigure}
    \caption{Real-world Data Results.}
    \label{fig:real}
\end{figure}

\paragraph{Experiment Setup.} We consider real-world datasets, including Lymphoma (biology) \citep{yuan2013truncated}, NUS-WIDE (web images) \citep{nus-wide-civr09}, and Numerical Numbers (handwritten numbers) \citep{asuncion2007uci}. We compare Go and IPU with SOAP, SRT, TPower (G) and report the results of TPower (D) as a baseline.
TPower (G) selects the sparsity pattern  with the leading eigenvector Greedily and TPower (D) uses the Deflation scheme, which cannot produce consistent sparsity pattern among rows.
We follow \citep{wang2014tighten} to use Convex Relaxation as the initialization. 
Following \citep{wang2014tighten,yang2015streaming}, we use the Normalized Explained Variance as the performance measure. The Normalized Explained Variance is defined as
$
\Tr(\widehat{\mat{W}}^\top \mat{A}\widehat{\mat{W}}) / \Tr(\mat{A}_m),
$
where the $\widehat{\mat{W}}$ is the subspace estimation returned by algorithms.

\paragraph{Results.} The experimental results are reported in \Cref{fig:real}, from which we get the following insights:
(1) For all three real-world datasets, the new algorithms, Go and IPU, consistently perform better than other state-of-the-art methods that solve FSPCA;
(2) For NN dataset, the performance of all methods except SOAP and TPower (D) are tied. It is of interest to see whether the reason for this phenomenon is the 
dataset is too difficult or too easy. Therefore, we plot the approximation bound in \Cref{thm:approx}, which reveals that these methods
achieve almost optimal performance;
(3) While TPower (D) achieves the highest NEV, it cannot be used for either feature selection or sparse subspace estimation (see \Cref{def:sparsespace}), due to the sparsity inconsistent issue of one-by-one eigenvectors estimation (see \Cref{fig:matrix}). Strictly speaking, TPower (D) actually solves a less constrained problem.

\begin{wrapfigure}{r}{0.33\textwidth}
\vspace{-1.em}
  \centering
    \includegraphics[width=0.33\textwidth]{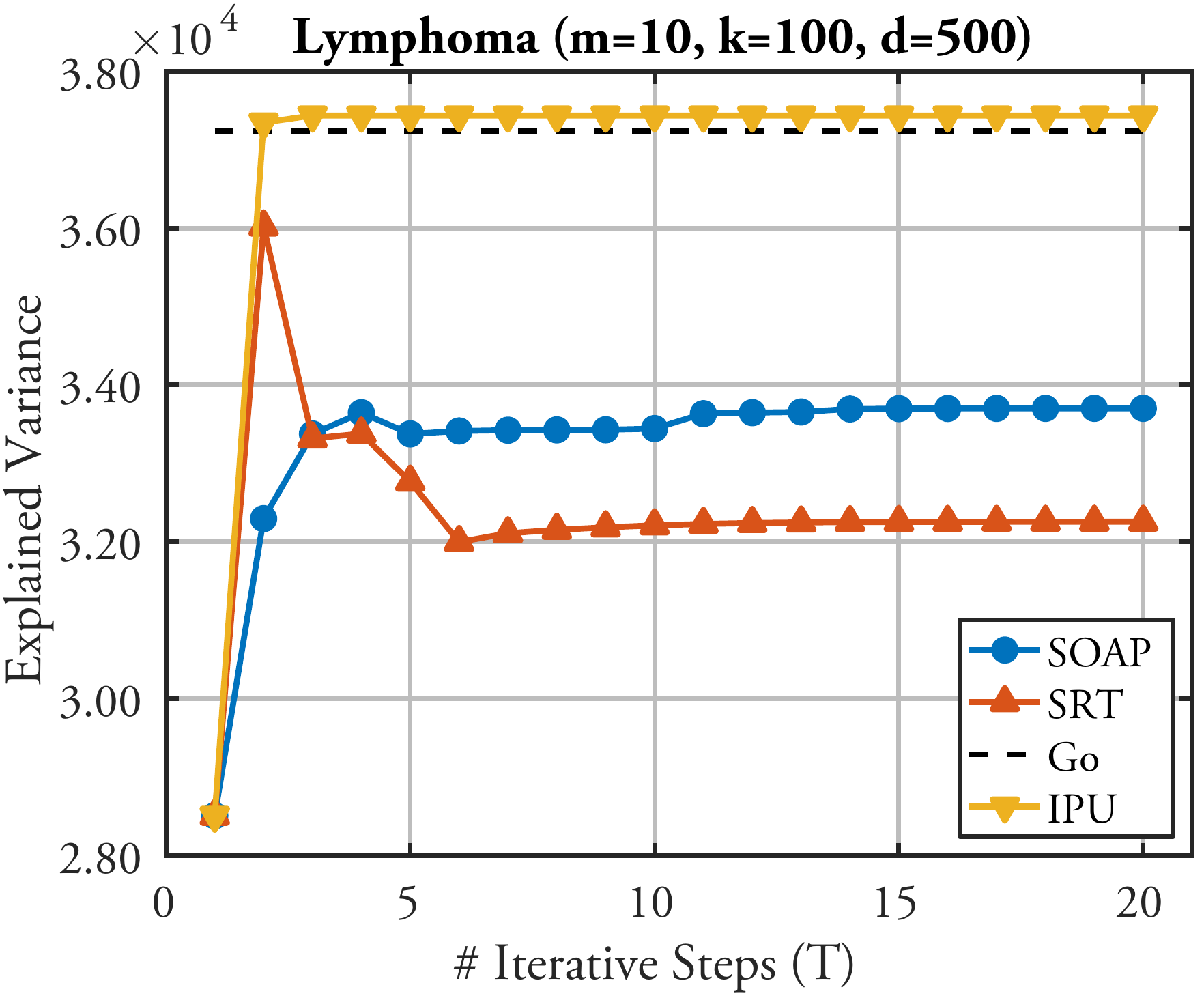}
 \caption{Convergence.}%
 \label{fig:convergence}%
 \vspace{-2em}
\end{wrapfigure}%

\paragraph{Convergence.}
In \Cref{thm:ascent}, we prove the monotonic ascent property of IPU (Algorithm \ref{alg:generalFSPCA}) and in \Cref{rem:ascent},
we claim that existing iterative schemes are not monotonic ascent guaranteed.
Here we provide numerical evidence to support this claim.
We run Go, IPU, SOAP, SRT on Lymphoma dataset with $m=10,k=100,d=500$.
We use the same convex relaxation initialization for all methods with row truncation.
We record the objective value in every iterative step for all methods.
The results are plotted in \Cref{fig:convergence}, from which we can see both SOAP and SRT are not ascent methods and
both Go and IPU achieve better Explained Variance than SOAP and SRT with the same initialization.
Besides, IPU takes less than 20 steps to converge, which is the case we keep seeing in all our experiments.

\section{Conclusion} \label{sec:conclusion}
In this paper, we present algorithms to directly estimate the row sparsity constrained leading $m$ eigenvectors.
We propose Algorithm \ref{alg:kISc} to solve FSPCA for low-rank covariance globally.
For general high-rank covariance, we propose Algorithm \ref{alg:generalFSPCA} to solve FSPCA by iteratively building a carefully designed low-rank proxy covariance matrix. 
We prove theoretical guarantees for both algorithms on approximation and convergence.
Experimental results show the promising performance of the new algorithms compared with the state-of-the-art methods.


\bibliography{l20}

\begin{thebibliography}{37}
\providecommand{\natexlab}[1]{#1}
\providecommand{\url}[1]{\texttt{#1}}
\expandafter\ifx\csname urlstyle\endcsname\relax
  \providecommand{\doi}[1]{doi: #1}\else
  \providecommand{\doi}{doi: \begingroup \urlstyle{rm}\Url}\fi

\bibitem[Asteris et~al.(2014)Asteris, Papailiopoulos, and
  Dimakis]{asteris2014nonnegative}
Megasthenis Asteris, Dimitris Papailiopoulos, and Alexandros Dimakis.
\newblock Nonnegative sparse pca with provable guarantees.
\newblock In \emph{International Conference on Machine Learning}, pages
  1728--1736, 2014.

\bibitem[Asuncion and Newman(2007)]{asuncion2007uci}
Arthur Asuncion and David Newman.
\newblock Uci machine learning repository, 2007.

\bibitem[Bertsekas and Tseng(1994)]{bertsekas1994partial}
Dimitri~P Bertsekas and Paul Tseng.
\newblock Partial proximal minimization algorithms for convex pprogramming.
\newblock \emph{SIAM Journal on Optimization}, 4\penalty0 (3):\penalty0
  551--572, 1994.

\bibitem[Breslau et~al.(1999)Breslau, Cao, Fan, Phillips, Shenker,
  et~al.]{breslau1999web}
Lee Breslau, Pei Cao, Li~Fan, Graham Phillips, Scott Shenker, et~al.
\newblock Web caching and zipf-like distributions: Evidence and implications.
\newblock In \emph{Ieee Infocom}, volume~1, pages 126--134. INSTITUTE OF
  ELECTRICAL ENGINEERS INC (IEEE), 1999.

\bibitem[Cai et~al.(2013)Cai, Nie, and Huang]{cai2013exact}
Xiao Cai, Feiping Nie, and Heng Huang.
\newblock Exact top-k feature selection via l2, 0-norm constraint.
\newblock In \emph{Twenty-Third International Joint Conference on Artificial
  Intelligence}, 2013.

\bibitem[Chan et~al.(2016)Chan, Papailliopoulos, and
  Rubinstein]{chan2016approximability}
Siu~On Chan, Dimitris Papailliopoulos, and Aviad Rubinstein.
\newblock On the approximability of sparse pca.
\newblock In \emph{Conference on Learning Theory}, pages 623--646, 2016.

\bibitem[Chua et~al.(July 8-10, 2009)Chua, Tang, Hong, Li, Luo, and
  Zheng]{nus-wide-civr09}
Tat-Seng Chua, Jinhui Tang, Richang Hong, Haojie Li, Zhiping Luo, and Yan-Tao
  Zheng.
\newblock Nus-wide: A real-world web image database from national university of
  singapore.
\newblock In \emph{Proc. of ACM Conf. on Image and Video Retrieval (CIVR'09)},
  Santorini, Greece., July 8-10, 2009.

\bibitem[d'Aspremont et~al.(2007)d'Aspremont, El~Ghaoui, Jordan, and
  Lanckriet]{d2007direct}
Alexandre d'Aspremont, Laurent El~Ghaoui, Michael~I. Jordan, and Gert R.~G.
  Lanckriet.
\newblock A direct formulation for sparse pca using semidefinite programming.
\newblock \emph{SIAM Rev.}, 49\penalty0 (3):\penalty0 434--448, July 2007.
\newblock ISSN 0036-1445.

\bibitem[Dempster et~al.(1977)Dempster, Laird, and Rubin]{dempster1977maximum}
Arthur~P Dempster, Nan~M Laird, and Donald~B Rubin.
\newblock Maximum likelihood from incomplete data via the em algorithm.
\newblock \emph{Journal of the Royal Statistical Society. Series B
  (methodological)}, pages 1--38, 1977.

\bibitem[Du et~al.(2018)Du, Nie, Wang, Yang, and Zhou]{du2018exploiting}
Xingzhong Du, Feiping Nie, Weiqing Wang, Yi~Yang, and Xiaofang Zhou.
\newblock Exploiting combination effect for unsupervised feature selection by
  $\ell_{2, 0}$ norm.
\newblock \emph{IEEE Trans. Neural Netw. Learn. Syst.}, \penalty0
  (99):\penalty0 1--14, 2018.

\bibitem[Faloutsos et~al.(1999)Faloutsos, Faloutsos, and
  Faloutsos]{faloutsos1999power}
Michalis Faloutsos, Petros Faloutsos, and Christos Faloutsos.
\newblock On power-law relationships of the internet topology.
\newblock In \emph{ACM SIGCOMM computer communication review}, volume~29, pages
  251--262. ACM, 1999.

\bibitem[Horn and Johnson(2012)]{horn2012matrix}
Roger~A Horn and Charles~R Johnson.
\newblock \emph{Matrix analysis}.
\newblock Cambridge university press, 2012.

\bibitem[Johnstone and Lu(2009)]{johnstone2009consistency}
Iain~M Johnstone and Arthur~Yu Lu.
\newblock On consistency and sparsity for principal components analysis in high
  dimensions.
\newblock \emph{Journal of the American Statistical Association}, 104\penalty0
  (486):\penalty0 682--693, 2009.

\bibitem[Kundu et~al.(2017)Kundu, Drineas, and
  Magdon-Ismail]{kundu2017recovering}
Abhisek Kundu, Petros Drineas, and Malik Magdon-Ismail.
\newblock Recovering pca and sparse pca via hybrid-(l 1, l 2) sparse sampling
  of data elements.
\newblock \emph{The Journal of Machine Learning Research}, 18\penalty0
  (1):\penalty0 2558--2591, 2017.

\bibitem[Lei et~al.(2015)Lei, Vu, et~al.]{lei2015sparsistency}
Jing Lei, Vincent~Q Vu, et~al.
\newblock Sparsistency and agnostic inference in sparse pca.
\newblock \emph{The Annals of Statistics}, 43\penalty0 (1):\penalty0 299--322,
  2015.

\bibitem[Lipp and Boyd(2016)]{lipp2016variations}
Thomas Lipp and Stephen Boyd.
\newblock Variations and extension of the convex--concave procedure.
\newblock \emph{Optimization and Engineering}, 17\penalty0 (2):\penalty0
  263--287, 2016.

\bibitem[Luss and d’Aspremont(2010)]{luss2010clustering}
Ronny Luss and Alexandre d’Aspremont.
\newblock Clustering and feature selection using sparse principal component
  analysis.
\newblock \emph{Optimization and Engineering}, 11\penalty0 (1):\penalty0
  145--157, 2010.

\bibitem[Mackey(2009)]{mackey2009deflation}
Lester~W Mackey.
\newblock Deflation methods for sparse pca.
\newblock In \emph{Advances in Neural Information Processing Systems}, pages
  1017--1024, 2009.

\bibitem[Magdon-Ismail and Boutsidis(2016)]{magdon2016optimal}
Malik Magdon-Ismail and Christos Boutsidis.
\newblock Optimal sparse linear encoders and sparse pca.
\newblock In \emph{Advances in Neural Information Processing Systems}, pages
  298--306, 2016.

\bibitem[Mihail and Papadimitriou(2002)]{mihail2002eigenvalue}
Milena Mihail and Christos Papadimitriou.
\newblock On the eigenvalue power law.
\newblock In \emph{International Workshop on Randomization and Approximation
  Techniques in Computer Science}, pages 254--262. Springer, 2002.

\bibitem[Moghaddam et~al.(2006)Moghaddam, Weiss, and
  Avidan]{moghaddam2006spectral}
Baback Moghaddam, Yair Weiss, and Shai Avidan.
\newblock Spectral bounds for sparse pca: Exact and greedy algorithms.
\newblock In \emph{Advances in neural information processing systems}, pages
  915--922, 2006.

\bibitem[Naikal et~al.(2011)Naikal, Yang, and Sastry]{naikal2011informative}
Nikhil Naikal, Allen~Y Yang, and S~Shankar Sastry.
\newblock Informative feature selection for object recognition via sparse pca.
\newblock In \emph{2011 International Conference on Computer Vision}, pages
  818--825. IEEE, 2011.

\bibitem[Pang et~al.(2018)Pang, Nie, Han, and Li]{pang2018efficient}
Tianji Pang, Feiping Nie, Junwei Han, and Xuelong Li.
\newblock Efficient feature selection via $ l_{2, 0}$-norm constrained sparse
  regression.
\newblock \emph{IEEE Transactions on Knowledge and Data Engineering}, 2018.

\bibitem[Papailiopoulos et~al.(2013)Papailiopoulos, Dimakis, and
  Korokythakis]{papailiopoulos2013sparse}
Dimitris Papailiopoulos, Alexandros Dimakis, and Stavros Korokythakis.
\newblock Sparse pca through low-rank approximations.
\newblock In \emph{International Conference on Machine Learning}, pages
  747--755, 2013.

\bibitem[Parikh et~al.(2014)Parikh, Boyd, et~al.]{parikh2014proximal}
Neal Parikh, Stephen Boyd, et~al.
\newblock Proximal algorithms.
\newblock \emph{Foundations and Trends{\textregistered} in Optimization},
  1\penalty0 (3):\penalty0 127--239, 2014.

\bibitem[Shen and Huang(2008)]{shen2008sparse}
Haipeng Shen and Jianhua~Z Huang.
\newblock Sparse principal component analysis via regularized low rank matrix
  approximation.
\newblock \emph{Journal of Multivariate Analysis}, 99\penalty0 (6):\penalty0
  1015--1034, 2008.

\bibitem[Sun et~al.(2017)Sun, Babu, and Palomar]{sun2017majorization}
Ying Sun, Prabhu Babu, and Daniel~P Palomar.
\newblock Majorization-minimization algorithms in signal processing,
  communications, and machine learning.
\newblock \emph{IEEE Transactions on Signal Processing}, 65\penalty0
  (3):\penalty0 794--816, 2017.

\bibitem[Von~Neumann(1937)]{von1937some}
John Von~Neumann.
\newblock \emph{Some matrix-inequalities and metrization of matric space}.
\newblock 1937.

\bibitem[Vu et~al.(2013{\natexlab{a}})Vu, Cho, Lei, and Rohe]{vu2013fantope}
Vincent~Q Vu, Juhee Cho, Jing Lei, and Karl Rohe.
\newblock Fantope projection and selection: A near-optimal convex relaxation of
  sparse pca.
\newblock In \emph{Advances in Neural Information Processing Systems}, pages
  2670--2678, 2013{\natexlab{a}}.

\bibitem[Vu et~al.(2013{\natexlab{b}})Vu, Lei, et~al.]{vu2013minimax}
Vincent~Q Vu, Jing Lei, et~al.
\newblock Minimax sparse principal subspace estimation in high dimensions.
\newblock \emph{The Annals of Statistics}, 41\penalty0 (6):\penalty0
  2905--2947, 2013{\natexlab{b}}.

\bibitem[Wang et~al.(2014)Wang, Lu, and Liu]{wang2014tighten}
Zhaoran Wang, Huanran Lu, and Han Liu.
\newblock Tighten after relax: Minimax-optimal sparse pca in polynomial time.
\newblock In \emph{Advances in neural information processing systems}, pages
  3383--3391, 2014.

\bibitem[Yang et~al.(2016)Yang, Ma, and Buja]{yang2016rate}
Dan Yang, Zongming Ma, and Andreas Buja.
\newblock Rate optimal denoising of simultaneously sparse and low rank
  matrices.
\newblock \emph{The Journal of Machine Learning Research}, 17\penalty0
  (1):\penalty0 3163--3189, 2016.

\bibitem[Yang and Xu(2015)]{yang2015streaming}
Wenzhuo Yang and Huan Xu.
\newblock Streaming sparse principal component analysis.
\newblock In \emph{International Conference on Machine Learning}, pages
  494--503, 2015.

\bibitem[Yuan and Zhang(2013)]{yuan2013truncated}
Xiao-Tong Yuan and Tong Zhang.
\newblock Truncated power method for sparse eigenvalue problems.
\newblock \emph{Journal of Machine Learning Research}, 14\penalty0
  (Apr):\penalty0 899--925, 2013.

\bibitem[Yuille and Rangarajan(2002)]{yuille2002concave}
Alan~L Yuille and Anand Rangarajan.
\newblock The concave-convex procedure (cccp).
\newblock In \emph{Advances in Neural Information Processing Systems}, pages
  1033--1040, 2002.

\bibitem[Zhang and Han(2018)]{zhang2018optimal}
Anru Zhang and Rungang Han.
\newblock Optimal sparse singular value decomposition for high-dimensional
  high-order data.
\newblock \emph{Journal of the American Statistical Association}, pages 1--40,
  2018.

\bibitem[Zou et~al.(2006)Zou, Hastie, and Tibshirani]{zou2006sparse}
Hui Zou, Trevor Hastie, and Robert Tibshirani.
\newblock Sparse principal component analysis.
\newblock \emph{Journal of Computational and Graphical Statistics}, 15\penalty0
  (2):\penalty0 265--286, 2006.

\end{thebibliography}
\bibliographystyle{plainnat}

\newpage
\appendix

\section*{Appendix}
\section{Proof of \Cref{thm:globaloptimality}}

\begin{definition}[Set of $k$th order principal submatrices]
For $m \leq k \leq d$ and matrix $\mat{A} \in \mathbb{R}^{d\times d}$, 
we define the set of $k$th order principal submatrices of $\mat{A}$ as\footnote{We use $[d]$ for $\{1,2,...,d\}$.}
\[
\mathbb{M}_k(\mat{A})= \left\{ \mat{A}_{\mathcal{I},\mathcal{I}} : \mathcal{I} \subseteq [d], \mathcal{I}=\texttt{Sort}(\mathcal{I}), \text{card}(\mathcal{I})=k\ \right\}.
\]
\end{definition}


\globaloptimality*

\begin{proof}

We start with an interesting observation. 
When we set $k=m$ in Problem \eqref{eq:fspca} (do not require rank$(\mat{A}) \leq m$), we are asking for the best $m$ features for projecting the original data into 
the best fit $m$ dimensional subspace.
When features are independent, this setting seems reasonable. Specifically, the problem we are talking about is 
\[
\max_{\mat{W}^\top\mat{W}=\Ind{m}, \|\mat{W}\|_{2,0} \leq m} \Tr\left(\mat{W}^\top \mat{A} \mat{W}\right).
\]
Note that for each $\mat{W}^\top\mat{W}=\Ind{m}, \|\mat{W}\|_{2,0}\leq m$, we can rewrite it as $\mat{W}=\mat{S}\mat{V}$, where $\mat{V}\in\mathbb{R}^{m\times m}$ satisfies $\mat{V}^\top\mat{V}=\Ind{m}$ and the row selection matrix $\mat{S}\in\{0,1\}^{d\times m}$ satisfies $\mat{S}^\top\mathbb{1}_d = \mathbb{1}_m$. 
It is easy to verify, for given $\mat{A}\in\mathbb{R}^{d\times d}$, 
\[
\{\mat{S}^\top \mat{A} \mat{S} : \mat{S}=\mathbb{S}_{d,m}(\texttt{Sort}(\mathcal{I})),\mathcal{I} \subseteq [d], \text{card}(\mathcal{I})=m\ \} = \mathbb{M}_m(\mat{A}).
\]
Therefore, above problem is equivalent to
\begin{equation} \label{eq:k_is_c_A_in_M}
\max_{\substack{\mat{V}\in\mathbb{R}^{m\times m}, \mat{V}^\top\mat{V}=\Ind{m} \\
\wt{\mat{A}} \in \mathbb{M}_m (\mat{A}) } } \Tr\left(\mat{V}^\top \wt{\mat{A}} \mat{V}\right).
\end{equation}
Note that $\mat{V}^\top\mat{V} = \mat{V}\mat{V}^\top = \Ind{m}$ since $\mat{V}$ is square (which is not true when $k \neq m$). Combining with the fact $\Tr(\mat{V}^\top \wt{\mat{A}} \mat{V})=\Tr(\wt{\mat{A}} \mat{V}\mat{V}^\top)$, Problem \eqref{eq:k_is_c_A_in_M} can be rewritten as
\[
\max_{ \wt{\mat{A}} \in \mathbb{M}_m (\mat{A}) } \Tr\left(\wt{\mat{A}}\right),
\]
which can be solved globally by sorting and selecting the $k$ largest diagonal elements of $\mat{A}$.

If we consider above argument carefully, we will realize the key point is that by setting $k=m$, we are able to write $\sum_{i=1}^m \lambda_i(\mat{S}^\top \mat{A} \mat{S})$ as $\Tr(\mat{S}^\top \mat{A} \mat{S})$. Equivalently, rank$(\wt{\mat{A}}) = \textnormal{rank}(\mat{S}^\top \mat{A} \mat{S})\leq m$. 

Note that if rank$(\mat{A}) \leq m$, then for all $k$ satisfies $m \leq k \leq d$, we have rank$(\wt{\mat{A}})\leq m$ where $\wt{\mat{A}} \in \mathbb{M}_k(\mat{A})$. Thus, if rank$(\mat{A}) \leq m$, we can use the same technique to solve the following problem even if $k \neq m$:
\begin{equation}\label{eq:general_global}
\max_{\substack{\mat{W}^\top\mat{W}=\Ind{m}, \|\mat{W}\|_{2,0} \leq k \\ \textnormal{rank}(\mat{A}) \leq m } } \Tr\left(\mat{W}^\top \mat{A} \mat{W}\right).
\end{equation}
In detail, note that
\[
\text{Prob. }\eqref{eq:general_global}
\Leftrightarrow \max_{\wt{\mat{A}} \in \mathbb{M}_k (\mat{A})} \sum_{i=1}^k \lambda_i(\wt{\mat{A}}) 
\Leftrightarrow \max_{\wt{\mat{A}} \in \mathbb{M}_k (\mat{A})} \Tr\left(\wt{\mat{A}}\right),
\]
which can be easily solved globally by first sorting the diagonal elements of $\mat{A}$ and selecting the $k$ largest elements then performing eigenvalue decomposition on the selected principal submatrix of $\mat{A}$ to obtain $\mat{W}$.
\end{proof}

\section{Proof of Convergence Guarantee}

Before proving the ascent theorem, we first introduce some preliminary results.

\begin{lemma}[\citealt{horn2012matrix}, Theorem 1.3.22]\label{lem:ABBA}
For $\mat{A}\in\mathbb{R}^{n_1\times n_2}, \mat{B}\in\mathbb{R}^{n_2\times n_1}$ with $n_1\leq n_2$, we have
\[
\lambda_i(\mat{BA}) = 
\left\{ \begin{array}{rcl}
         \lambda_{i}(\mat{AB}) & \mbox{for} & 1\leq i \leq n_1 \\ 
         0 & \mbox{for} & n_1+1\leq i \leq n_2.
\end{array}
\right.
\]
\end{lemma}

\Cref{lem:ABBA} leads to an eigenvalue estimation that will be used in our main proof.
\begin{corollary}\label{coro:eigenAB}
Let $\mat{\Gamma} = \mat{X}^\top \mat{W}_{t}(\mat{W}_{t}^\top\mat{XX}^\top\mat{W}_{t})^{\dagger}\mat{W}_{t}^\top\mat{X}$. For the eigenvalues of $\mat{\Gamma}$, it holds
\[
\lambda_i(\mat{\Gamma}) = 
\left\{ \begin{array}{rcl}
         1 & \mbox{for} & 1\leq i \leq r \\ 
         0 & \mbox{for} & r+1\leq i \leq d,
\end{array}
\right.
\]
where $r = \textnormal{rank}(\mat{X}^\top \mat{W}_t) \leq m$.
\end{corollary}

\begin{proof}
Let $\mat{A} = (\mat{W}_{t}^\top\mat{XX}^\top\mat{W}_{t})^{\dagger}\mat{W}_{t}^\top\mat{X}, \mat{B} = \mat{X}^\top \mat{W}_{t}$.
Thus, for each $1\leq i \leq d$, $\lambda_i(\mat{\Gamma}) = \lambda_i(\mat{BA}) $ and 
\[
\mat{AB} = (\mat{W}_{t}^\top\mat{XX}^\top\mat{W}_{t})^{\dagger}\mat{W}_{t}^\top\mat{X}\mat{X}^\top \mat{W}_{t}.
\]
Using \Cref{lem:ABBA}, we have
\[
\lambda_i(\mat{\Gamma}) = 
\left\{ \begin{array}{rcl}
         \lambda_{i}(\mat{AB}) & \mbox{for} & 1\leq i \leq m \\ 
         0 & \mbox{for} & m+1\leq i \leq d.
\end{array}
\right.
\]
Note that $\textnormal{rank}(\mat{AB}) = r \leq m$ and
\[
\lambda_i(\mat{AB}) = 
\left\{ \begin{array}{rcl}
         1 & \mbox{for} & 1\leq i \leq r \\ 
         0 & \mbox{for} & r+1\leq i \leq d.
\end{array}
\right.
\]
which completes the proof.
\end{proof}

\begin{lemma}[\citealt{von1937some}]\label{lem:von}
If matrices $\mat{X} \in \mathbb{R}^{n\times n}$ and $\mat{Y} \in \mathbb{R}^{n\times n}$ are symmetric,
then,
\[
\Tr(\mat{X}\mat{Y}) \leq \sum_{i=1}^n \lambda_i(\mat{X})\lambda_i(\mat{Y}).
\]
\end{lemma}

Now we are ready to prove the main result which shows the objective function values generated by Algorithm \ref{alg:generalFSPCA} are monotonic ascent.

\ascentthm*
\begin{proof}
Note that
\[
\begin{aligned}& \Tr\left(\mat{W}_{t}^\top\mat{A}\mat{W}_{t} \right) \\ 
\overset{\text{\ding{172}}}{=} & \Tr\left(\mat{W}_{t}^\top\mat{A}\mat{W}_{t}(\mat{W}_{t}^\top\mat{AW}_{t})^{\dagger}\mat{W}_{t}^\top\mat{AW}_{t}\right)\\
\overset{\text{\ding{173}}}{\leq} & \Tr\left(\wt{\mat{W}}_{t+1}^\top\mat{A}\mat{W}_{t}(\mat{W}_{t}^\top\mat{AW}_{t})^{\dagger}\mat{W}_{t}^\top\mat{A}\wt{\mat{W}}_{t+1}\right)\\
\overset{\text{\ding{174}}}{=} & \Tr\left(\mat{X}^\top\mat{W}_{t}(\mat{W}_{t}^\top\mat{AW}_{t})^{\dagger}\mat{W}_{t}^\top\mat{X}\mat{X}^\top\wt{\mat{W}}_{t+1}\wt{\mat{W}}_{t+1}^\top\mat{X} \right) \label{eq:eqeqeq}
\end{aligned}
\]
where 
\ding{172} uses the fact $\mat{A}=\mat{AA}^{\dagger}\mat{A}$;
\ding{173} uses $\wt{\mat{W}}_{t+1}$ maximizing Problem \eqref{eq:fspca} for $\mat{P}_t$; 
\ding{174} uses $\mat{A}\succeq \mathbb{0}$, which implies that we can always find $\mat{X}\in\mathbb{R}^{d\times d}$ such that $\mat{A}=\mat{X}\mat{X}^\top$ (e.g., with Cholesky decomposition).

Let $\mat{\Gamma}\in\mathbb{R}^{d\times d}, \mat{\Omega}\in\mathbb{R}^{d\times d}$ be 
\[
\begin{aligned}
\mat{\Gamma} =& \mat{X}^\top\mat{W}_{t}(\mat{W}_{t}^\top\mat{AW}_{t})^{\dagger}\mat{W}_{t}^\top\mat{X} \\
\mat{\Omega} =& \mat{X}^\top\wt{\mat{W}}_{t+1}\wt{\mat{W}}_{t+1}^\top\mat{X}.
\end{aligned}
\]
Then, the RHS (right-hand side) of \ding{174} can be rewritten as
\[
\text{RHS of \ding{174}} =  \Tr(\mat{\Gamma\Omega}) 
\overset{\text{\ding{175}}}{\leq}  \sum_{i=1}^{d}\lambda_{i}(\mat{\Gamma})\lambda_{i}(\mat{\Omega})
\overset{\text{\ding{176}}}{\leq}  \sum_{i=1}^{m}\lambda_{i}(\mat{\Omega}),
\]
where 
\ding{175} uses \Cref{lem:von};
\ding{176} uses \Cref{coro:eigenAB} and the fact for each $1\leq i \leq m$, we have $\lambda_i(\mat{\Omega}) \geq 0$.

Note that $\textnormal{rank}(\mat{\Omega}) \leq \textnormal{rank}(\wt{\mat{W}}_{t+1}) = m$. Then we have $\sum_{i=1}^{m}\lambda_{i}(\mat{\Omega}) = \Tr(\mat{\Omega})$. Thus the RHS of \ding{176} can be rewritten as
\[
\text{RHS of \ding{176}}
= \Tr(\mat{\Omega}) = \Tr(\wt{\mat{W}}_{t+1}^\top\mat{A}\wt{\mat{W}}_{t+1}),
\]
which is exactly the updated objective function value of Problem \eqref{eq:WwtDEF}. 
But we can go further by notice that $\wt{\mat{W}}_{t+1} = \mat{S}_{t+1}\wt{\mat{V}}_{t+1}$
, $\mat{W}_{t+1} = \mat{S}_{t+1}\mat{V}_{t+1}$, and $\mat{V}_{t+1}$ maximizes Problem \eqref{eq:solveVclever}.
That gives
\[
\Tr(\wt{\mat{W}}_{t+1}^\top\mat{A}\wt{\mat{W}}_{t+1}) 
=  \Tr(\wt{\mat{V}}_{t+1}^\top \mat{S}_{t+1}^\top \mat{A} \mat{S}_{t+1} \wt{\mat{V}}_{t+1}) \\
\leq  \Tr(\mat{W}_{t+1}^\top\mat{A}\mat{W}_{t+1}),
\]
which completes the proof.
\end{proof}

\section{Proof of \Cref{chaim:PisSolvable}}

\proxyclaim*
\begin{proof}
The first part is from $\textnormal{rank}(\mat{P}_t) \leq \textnormal{rank}(\mat{W}_t) = m$.
Let $\mat{\Phi} = \mat{A}\mat{W}_{t}(\mat{W}^\top_t \mat{A}\mat{W}_t)^{\dagger}\mat{W}_{t}^\top \mat{X}$.
Using the facts $\mat{A}\succeq \mathbb{0}$ (which implies $\mat{A}=\mat{XX}^\top$ with Cholesky) and $\mat{B}^\dagger=\mat{B}^\dagger\mat{B} \mat{B}^\dagger$, the second part is from 
\[
\mat{P}_t = \mat{A}\mat{W}_{t}(\mat{W}^\top_t \mat{A}\mat{W}_t)^{\dagger}\mat{W}_{t}^\top \mat{X}\mat{X}^\top \mat{W}_t (\mat{W}^\top_t \mat{A}\mat{W}_t)^{\dagger} \mat{W}_t^\top \mat{A} = \mat{\Phi\Phi}^\top \succeq 0,
\]
which completes the proof.
\end{proof}

\section{Proof of \Cref{claim:invert}}

First of all, we need a result to bound the eigenvalues of principal submatrix.

\begin{lemma}[\citealt{horn2012matrix}, Theorem 4.3.28]\label{lem:interlacing}
Let $\mat{A} \in \mathbb{R}^{d\times d}$ be symmetric matrix that can be partitioned as
\[
\mat{A} = 
\left[ \begin{array}{cc}
     \mat{B} & \mat{C} \\ \mat{C}^\top & \mat{D} \end{array}
\right],
\]
where $\mat{B} \in \mathbb{R}^{k\times k},
\mat{C} \in \mathbb{R}^{k\times (d-k)},
\mat{D} \in \mathbb{R}^{(d-k)\times(d-k)}$.
Let the eigenvalues of $\mat{A}$ and $\mat{B}$ be sorted in descending order. Then, for each $1\leq i \leq k$, we have
$
\lambda_i(\mat{A}) \geq \lambda_i(\mat{B}) \geq \lambda_{d-k+i}(\mat{A}).
$
\end{lemma}

\invertclaim*
\begin{proof}
For ease of notation, we denote $\mat{W}_t$ by $\mat{W}$. 
Recall we can always extend the semi-orthonormal matrix $\mat{W}$ to an orthonormal matrix $\overline{\mat{W}} = [\mat{W} \ \ \mat{W}_{\bot}]$.
We can write $\mat{W}^\top\mat{A}\mat{W}$ as a block since
\[
\overline{\mat{W}}^\top \mat{A} \overline{\mat{W}} = 
\left[ 
\begin{array}{cc}
     \mat{W}^\top \mat{A} \mat{W} & \mat{W}^\top\mat{A}\mat{W}_{\bot} \\ 
     \mat{W}^\top_\bot\mat{A}\mat{W} & \mat{W}^\top_{\bot}\mat{A}\mat{W}_{\bot} 
\end{array}
\right].
\]
Note that for each $1\leq i \leq d$, we have $\lambda_i(\overline{\mat{W}}^\top \mat{A} \overline{\mat{W}}) = \lambda_i(\mat{A})$ since $\overline{\mat{W}}$ is orthonormal.
Using \Cref{lem:interlacing}, we have for each $1\leq i \leq m$,
\[
\lambda_i(\mat{W}_t^\top\mat{A}\mat{W}_t) \geq \lambda_{d-k+i}(\overline{\mat{W}}^\top \mat{A} \overline{\mat{W}}) = \lambda_{d-k+i}(\mat{A}).
\]
Using rank$(\mat{A}) \geq d - k + m$, the proof completes.
\end{proof}

\section{Proof of Approximation Guarantee}
\approxgua*
\begin{proof}
Let $\mat{A}_m^c = \mat{A}-\mat{A}_m$.
Note that
\[
\begin{aligned}
&\Tr(\mat{W}_*^\top \mat{A}\mat{W}_*) \\
=&\max_{\mat{W}^\top\mat{W}=\Ind{m},\|\mat{W}\|_{2,0} \leq k} \Tr(\mat{W}^\top \mat{A}\mat{W}) \\ = & \max_{\mat{W}^\top\mat{W}=\Ind{m},\|\mat{W}\|_{2,0} \leq k} \Tr(\mat{W}^\top \mat{A}_m\mat{W}) + \Tr(\mat{W}^\top \mat{A}_m^c\mat{W}) \\
\leq & \max_{\mat{W}^\top\mat{W}=\Ind{m},\|\mat{W}\|_{2,0} \leq k} \Tr(\mat{W}^\top \mat{A}_m\mat{W}) + \max_{\mat{W}^\top\mat{W}=\Ind{m},\|\mat{W}\|_{2,0} \leq k}\Tr(\mat{W}^\top \mat{A}_m^c\mat{W}) \\
\leq & \Tr(\mat{W}_m^\top \mat{A}_m\mat{W}_m) + \max_{\mat{W}^\top\mat{W}=\Ind{m}}\Tr(\mat{W}^\top \mat{A}_m^c\mat{W}) \\
\leq & \Tr(\mat{W}_m^\top \mat{A}_m\mat{W}_m) + \sum_{i=m+1}^{2m} \lambda_i(\mat{A}),
\end{aligned}
\]
which indicates
\[
\frac{\Tr(\mat{W}_m^\top \mat{A}_m\mat{W}_m)}{\Tr(\mat{W}_*^\top \mat{A}\mat{W}_*)} \geq 1-\frac{\sum_{i=m+1}^{2m} \lambda_i(\mat{A})}{\Tr(\mat{W}_*^\top \mat{A}\mat{W}_*)}.
\]
Note that,
\[
\Tr(\mat{W}_*^\top \mat{A}\mat{W}_*) \geq \max_{\mat{W}^\top\mat{W}=\Ind{m},\|\mat{W}\|_{2,0} \leq m} \Tr(\mat{W}^\top \mat{A}\mat{W}) \geq \frac{m}{d}\Tr(\mat{A}) =  \frac{m}{d}\sum_{i=1}^d \lambda_i(\mat{A}),
\]
where the first inequality uses $m\leq k$ and the second inequality uses \Cref{thm:globaloptimality}.

Besides,
\[
\Tr(\mat{W}_*^\top \mat{A}\mat{W}_*) \geq \max_{\mat{W}^\top\mat{W}=\Ind{m},\|\mat{W}\|_{2,0} \leq k} \Tr(\mat{W}^\top \mat{A}_m\mat{W}) \geq \frac{k}{d}\sum_{i=1}^d \lambda_i(\mat{A}_m) = \frac{k}{d}\sum_{i=1}^m \lambda_i(\mat{A}),
\]
where the first inequality uses $\mat{A}\succeq \mathbb{0}$ and the second inequality uses \Cref{thm:globaloptimality}.

Let $r=\min\{\textnormal{rank}(\mat{A}), 2m\}$, and
\[
G_1 = \frac{\sum_{i=m+1}^{r}\lambda_i(\mat{A})}{\sum_{i=1}^m \lambda_i(\mat{A})},\qquad\qquad
G_2 = \frac{\sum_{i=m+1}^{r}\lambda_i(\mat{A})}{\sum_{i=1}^d \lambda_i(\mat{A})}.
\]
Thus, we have
\[
1 \geq \frac{\Tr(\mat{W}_m^\top \mat{A}_m\mat{W}_m)}{\Tr(\mat{W}_*^\top \mat{A}\mat{W}_*)} \geq 1-\min\left\{ \frac{d}{k}\cdot G_1,
\frac{d}{m}\cdot G_2
 \right\}.
\]
Because $\mat{A}\succeq \mathbb{0}$, we have
\[
\Tr(\mat{W}_m^\top\mat{A}\mat{W}_m) = \Tr(\mat{W}_m^\top\mat{A}_m\mat{W}_m)+\Tr(\mat{W}_m^\top\mat{A}_m^c\mat{W}_m) \geq \Tr(\mat{W}_m^\top\mat{A}_m\mat{W}_m),
\]
which completes the proof.
\end{proof} 

\section{Proof of Zipf-like \Cref{coro:power}}
To prove the corollary, we need the following auxiliary lemmas.

\begin{lemma}\label{lem:intEst}
\[
\frac{(b+1)^{1-t}-a^{1-t}}{1-t} \leq \sum_{i=a}^b \frac{1}{i^t} \leq \frac{b^{1-t}-(a-1)^{1-t}}{1-t}.
\]
\end{lemma}
\begin{proof} 
Using approximation by definite integrals, we have
\[
\begin{aligned}
 \sum_{i=a}^b \frac{1}{i^t} &\leq 
 \int_{a-1}^{b} \frac{1}{i^t} \text{d}i=
 \left.\frac{i^{1-t}}{1-t} \right|_{a-1}^{b}=
\frac{b^{1-t}-(a-1)^{1-t}}{1-t},\\
\sum_{i=a}^b \frac{1}{i^t} &\geq
\int_a^{b+1} \frac{1}{i^t} \text{d}i =
\left.\frac{i^{1-t}}{1-t} \right|_a^{b+1} =
\frac{(b+1)^{1-t}-a^{1-t}}{1-t}
\end{aligned}
\]
which completes the proof.
\end{proof}
\begin{lemma}\label{lem:SumDescent}
\[
\frac{\sum_{i=m+1}^{2m} \frac{1}{i^t}}{\sum_{i=1}^{m} \frac{1}{i^t}} \leq \frac{1}{2^t}.
\]
\end{lemma}
\begin{proof}
Let $m' = m + n$ with $n \geq 1$.
Note that
\[
\begin{aligned}
\sum_{i=m+1}^{m+n} \frac{1}{i^t} - \sum_{i=2m+1}^{2m+n} \frac{1}{i^t} &= \sum_{j=1}^n \left(\frac{1}{(m+j)^t} - \frac{1}{(2m+j)^t}\right) \geq 0 \\
\sum_{i=m+1}^{m+n} \frac{1}{i^t} - \sum_{i=2m+n+1}^{2m+2n} \frac{1}{i^t} &= \sum_{j=1}^n \left(\frac{1}{(m+j)^t} - \frac{1}{(2m+n+j)^t}\right) \geq 0, \\
\end{aligned}
\]
which implies
\[
\sum_{i=2m+1}^{2m'}\frac{1}{i^t} -\sum_{i=m+1}^{m'}\frac{1}{i^t} = 
\sum_{i=2m+1}^{2m+2k}\frac{1}{i^t} -\sum_{i=m+1}^{m+k}\frac{1}{i^t}=
\sum_{i=2m+1}^{2m+n} \frac{1}{i^t} + \sum_{i=2m+n+1}^{2m+2n} \frac{1}{i^t} -\sum_{i=m+1}^{m+k}\frac{1}{i^t}
\leq \sum_{i=m+1}^{m+n} \frac{1}{i^t}.
\]
Thus, we have
\[
\begin{aligned}
\frac{\displaystyle \sum_{i=m'+1}^{2m'} \frac{1}{i^t}}{\displaystyle \sum_{i=1}^{m'} \frac{1}{i^t}} &= 
\frac{\displaystyle \sum_{i=m+1}^{2m} \frac{1}{i^t}-\sum_{i=m+1}^{m'}\frac{1}{i^t} + \sum_{i=2m+1}^{2m'}\frac{1}{i^t} }{\displaystyle \sum_{i=1}^{m} \frac{1}{i^t}+\sum_{i=m+1}^{m'}\frac{1}{i^t} } \leq \frac{\displaystyle \sum_{i=m+1}^{2m} \frac{1}{i^t}+\sum_{i=m+1}^{m'}\frac{1}{i^t} }{\displaystyle \sum_{i=1}^{m} \frac{1}{i^t}+\sum_{i=m+1}^{m'}\frac{1}{i^t} } \leq 
\frac{\displaystyle \sum_{i=m+1}^{2m} \frac{1}{i^t} }{\displaystyle \sum_{i=1}^{m} \frac{1}{i^t} },
\end{aligned}
\]
where the last inequality uses the fact: if $0<a<b$ and $c >0$, then $\frac{a}{b} \leq \frac{a+c}{b+c}$.

Therefore, using $m\geq 1$, we have
\[
\frac{\sum_{i=m+1}^{2m} \frac{1}{i^t}}{\sum_{i=1}^{m} \frac{1}{i^t}} \leq \frac{\sum_{i=2}^{2} \frac{1}{i^t}}{\sum_{i=1}^{1} \frac{1}{i^t}} = \frac{1}{2^t},
\]
which completes the proof.
\end{proof}

\begin{lemma}[Bound $G_1$]\label{lem:coroEst}
\[
G_1 = \frac{\sum_{i=m+1}^{r}\lambda_i(\mat{A})}{\sum_{i=1}^m \lambda_i(\mat{A})} = 
\frac{\sum_{i=m+1}^{2m} \frac{1}{i^t}}{\sum_{i=1}^{m} \frac{1}{i^t}} \leq
\min\left\{ \frac{1}{m^{t-1}}, \frac{1}{2^{t}}\right\}.
\]
\end{lemma}
\begin{proof}
Using \Cref{lem:SumDescent}, we have
\[
\frac{\sum_{i=m+1}^{2m} \frac{1}{i^t}}{\sum_{i=1}^{m} \frac{1}{i^t}} \leq \frac{1}{2^t}.
\]
Leveraging the \Cref{lem:intEst}, we have
\[
\sum_{i=m+1}^{2m} \frac{1}{i^t} \leq \frac{m^{1-t}-(2m)^{1-t}}{t-1}, \quad 
\sum_{i=1}^{m} \frac{1}{i^t} \geq \frac{1-(m+1)^{1-t}}{t-1},
\]
which gives
\[
\frac{\sum_{i=m+1}^{2m} \frac{1}{i^t}}{\sum_{i=1}^{m} \frac{1}{i^t}} \leq
\frac{m^{1-t}-(2m)^{1-t}}{1-(m+1)^{1-t}} =
\frac{\frac{1}{m^{t-1}} \left(1-\frac{1}{2^{t-1}}\right)}{1-\frac{1}{(m+1)^{t-1}}} =
\frac{1-\frac{1}{2^{t-1}}}{m^{t-1}\left(1-\frac{1}{(m+1)^{t-1}}\right)}.
\]
Note that $m \geq 1$, which implies
\[
\frac{\sum_{i=m+1}^{2m} \frac{1}{i^t}}{\sum_{i=1}^{m} \frac{1}{i^t}} \leq \frac{1-\frac{1}{2^{t-1}}}{m^{t-1}\left(1-\frac{1}{2^{t-1}}\right)} \leq \frac{1}{m^{t-1}}.
\]
The proof completes.
\end{proof}

\coroPower*
\begin{proof}
Using \Cref{lem:coroEst}, we have
\[
G_1 = \frac{\sum_{i=m+1}^{r}\lambda_i(\mat{A})}{\sum_{i=1}^m \lambda_i(\mat{A})}
\leq \frac{\sum_{i=m+1}^{2m} \frac{1}{i^t}}{\sum_{i=1}^{m} \frac{1}{i^t}} \leq
\min\left\{ \frac{1}{m^{t-1}}, \frac{1}{2^{t}}\right\}.
\]
Then, the conclusion directly follows from \Cref{thm:approx}.
\end{proof}

\section{Experiments Details}\label{sec:expDetail}

\subsection{Synthetic Data Details}

For Scheme A and B, they are the synthetic data used in \citep{wang2014tighten}.
But we trim them to fit our setting, that is $m=3,k=7,d=20$.
For Scheme C, we validate the correctness that Algorithm \ref{alg:kISc} globally solves Problem \eqref{eq:general_global}.
For Scheme D, we use it to see the performance comparison when the rank$(\mat{A})$ is strictly larger than $m$.
For Scheme E and F, we compare the performance when data are generated from known distribution rather than using the eigenvalues fixed covariance.

For A--D, we fix the eigenvalues and generate the eigenspace randomly following \citep{wang2014tighten}.
\subsection{Performance Measures}

\paragraph{Intersection Ratio.}
  \[
\frac{\text{card}\left( \{\text{estimated indices}\} \cap \{\text{optimal indices} \} \right)}{\text{sparsity }k}.
  \]
  The reason we use Intersection Ratio is that FSPCA performs feature selection and PCA simultaneously. The Intersection Ratio can
  measure the intersection between the indices return by algorithm and the optimal indices.
\paragraph{Relative Error.}
  \[
\frac{\Tr\left(\mat{W}^\top\mat{A}\mat{W}\right) - \Tr\left(\mat{W}^{*\top}\mat{A}\mat{W}^*\right)}{\Tr\left(\mat{W}^{*\top}\mat{A}\mat{W}^*\right)}.
  \]
\paragraph{Hit Frequency.}
  \[
\frac{1}{N} \sum_{i=1}^N \mathbb{1}\Big\{ \text{Relative Error} \leq 10^{-3} \Big\},
  \]%
  where $N$ is the number of repeated running. This measure shows the frequency of the algorithm approximately reach the global optimum.

\subsection{Computing Infrastructure}
All experiments in this paper were run on a MacBook Pro laptop with 2.3 GHz Intel Core i5 CPU and 16GB memory.

\end{document}